\documentclass[conference]{IEEEtran}
\usepackage{fancyhdr}

\usepackage{graphics,epsfig,graphicx,float,color}
\usepackage[margin=0pt,labelsep=space,font=small,labelfont=bf,textfont=it]{caption}
\usepackage{atbegshi}
\usepackage{fancyhdr}
\usepackage{textcomp}
\usepackage{algorithm}

\usepackage[noend]{algpseudocode}

\usepackage{verbatimbox}
\usepackage{etex}
\usepackage{amsmath,amssymb,amsbsy,amsfonts,latexsym}

\usepackage{multicol,multirow}
\usepackage{soul}
\usepackage{hhline}
\usepackage{paralist}
\usepackage{array}
\usepackage{marvosym}
\usepackage{url}
\usepackage{boxedminipage}
\usepackage{wrapfig}
\usepackage{multirow}
\usepackage{threeparttable}
\usepackage{pdfpages}
\usepackage{xspace}
\usepackage{colortbl}
\usepackage{tabularx,booktabs}
\usepackage{xcolor}
\usepackage{tikz} 
\usetikzlibrary{arrows} 
\usetikzlibrary{patterns}  
\usepackage{pgfplots} 
\usepackage[switch]{lineno}
\pgfplotsset{compat=1.3}

\usepackage{enumitem}
\usepackage[plainpages=false, colorlinks=true,
    citecolor=blue, filecolor=blue, linkcolor=blue,
    urlcolor=blue]{hyperref}
\usepackage{siunitx}
\usepackage{booktabs}
\usepackage{calligra}

\DeclareMathAlphabet{\mathcalligra}{T1}{calligra}{m}{n}

\definecolor{light-gray}{gray}{0.8}

\newcolumntype{R}{>{\columncolor{light-gray}}r}
\newcolumntype{L}{>{\columncolor{light-gray}}l}
\newcolumntype{C}{>{\columncolor{light-gray}}c}

\tikzset{zigzag/.style={decorate, decoration=zigzag}}

\usepackage{amsthm}
\usepackage[capitalize]{cleveref}

\newtheorem{theorem}{Theorem}
\newtheorem{definition}{Definition}
\newtheorem{lemma}[theorem]{Lemma}
\newtheorem{proposition}[theorem]{Proposition}

\usepackage{mathtools}
\DeclareMathSymbol{\mh}{\mathord}{operators}{`\-}

\algnewcommand\algorithmicforeach{\textbf{foreach}}
\algdef{S}[FOR]{ForEach}[1]{\algorithmicforeach\ #1\ \algorithmicdo}
\algdef{SE}[DOWHILE]{Do}{doWhile}{\algorithmicdo}[1]{\algorithmicwhile\ #1}%
\usetikzlibrary{matrix}
\usetikzlibrary{arrows}

\newcommand{\commentsymbol}{//}
\algrenewcommand\algorithmiccomment[1]{\hfill \commentsymbol{} #1}
\algnewcommand{\LineComment}[1]{\State \(\commentsymbol{}\) #1}
\usepackage{setspace}
\usetikzlibrary{positioning}

\definecolor{t_green}{rgb}{0.533333333,	0.847058824,	0.690196078}
\definecolor{t_yellow}{rgb}{1,	0.964705882,	0}
\definecolor{t_orange}{rgb}{1,	0.8,	0.360784314}
\definecolor{t_blue}{rgb}{0.607843137,	0.866666667,	1}
\definecolor{t_red}{rgb}{1,0.435294118,0.411764706}

\usepackage{times,helvet,courier,xcolor}
\usepackage{tabularx,ragged2e,booktabs,caption}
\newcolumntype{C}{>{\Centering\arraybackslash}X} %

\renewcommand{\figurename}{Figure}
\renewcommand{\tablename}{Table}
\usepackage{textcmds}

\usetikzlibrary[pgfplots.groupplots,calc]

\pgfplotsset{grid style={dotted,black}}
\pgfplotsset{major grid style={dotted,black}}

\pgfplotsset{grid style={dotted,black}}
\pgfplotsset{major grid style={dotted,black}}
\usepackage{nameref}
\usepackage{subcaption}
\newcounter{mylabelcounter}

\makeatletter
\newcommand{\labelText}[2]{%
#1\refstepcounter{mylabelcounter}%
\immediate\write\@auxout{%
  \string\newlabel{#2}{{1}{\thepage}{{\unexpanded{#1}}}{mylabelcounter.\number\value{mylabelcounter}}{}}%
}%
}
\makeatother

\usepackage{bm}

\usepackage{diagbox}
\usepackage{makecell}
\usepackage{slashbox,pict2e}
\usepackage[textsize=tiny]{todonotes}

\makeatletter
\newcommand{\crefnames}[3]{%
  \@for\next:=#1\do{%
    \expandafter\crefname\expandafter{\next}{#2}{#3}%
  }%
}
\makeatother
\crefnames{part,chapter,section}{\S}{\S\S}

\usepackage{pifont}
\definecolor{LightCyan}{rgb}{0.88,1,1}
\renewcommand{\b}[1]{{\bf #1}}

\DeclareMathOperator*{\argmin}{arg\,min}
\DeclareMathOperator*{\argmax}{arg\,max}
\DeclareMathOperator{\Tr}{Trace}

\newcommand{\nbone}{\ding{182}\xspace}
\newcommand{\nbtwo}{\ding{183}\xspace}
\newcommand{\nbthree}{\ding{184}\xspace}
\newcommand{\nbfour}{\ding{185}\xspace}

\newcommand{\nfiral}{Approx-FIRAL\xspace}
\newcommand{\exactfiral}{Exact-FIRAL\xspace}

\newcommand{\Relax}{{\sc Relax}\xspace}
\newcommand{\Round}{{\sc Round}\xspace}

\newcommand{\poolacc}{pool accuracy\xspace}
\newcommand{\evalacc}{evaluation accuracy\xspace}

\renewcommand{\b}[1]{{\bf #1}}

\newcommand{\td}{\widetilde{d}}

\newcommand{\A}{\b{A}}
\newcommand{\B}{\b{B}}

\newcommand{\Xo}{\b{X}_o}
\newcommand{\Xu}{\b{X}_u}

\newcommand{\bH}{\b{H}}
\newcommand{\Ho}{\b{H}_o}
\newcommand{\tHo}{\widetilde{\b{H}}_o}

\newcommand{\Hp}{\b{H}_p}
\newcommand{\Hz}{\b{H}_z}
\newcommand{\Hi}{\b{H}_i}
\newcommand{\tHi}{\widetilde{\b{H}}_i}

\newcommand{\tH}{\widetilde{\b{H}}}

\newcommand{\bSigma}{\bm{\Sigma}}
\newcommand{\bSigmahalfinv}{{\bm{\Sigma}}_{\diamond}^{-1/2}}

\newcommand{\bLambda}{\bm{\Lambda}}

\let\subparagraph\paragraph
\usepackage{titlesec}
\let\subparagraph\llncssubparagraph
\titlespacing{\section}{5pt}{5pt}{5pt}
\titlespacing{\subsection}{5pt}{5pt}{5pt}

\def\BibTeX{{\rm B\kern-.05em{\sc i\kern-.025em b}\kern-.08em
    T\kern-.1667em\lower.7ex\hbox{E}\kern-.125emX}}

\newcolumntype{C}[1]{>{\centering\let\newline\\\arraybackslash\hspace{0pt}}m{#1}}

\begin{document}

\title{A Scalable Algorithm for Active Learning}

\author{\IEEEauthorblockN{Youguang Chen}
\IEEEauthorblockA{
\textit{University of Texas at Austin}\\
Austin, USA \\
youguang@utexas.edu} 
\and
\IEEEauthorblockN{Zheyu Wen}
\IEEEauthorblockA{
\textit{University of Texas at Austin}\\
Austin, USA \\
zheyw@utexas.edu}  
\and
\IEEEauthorblockN{George Biros}
\IEEEauthorblockA{
\textit{University of Texas at Austin}\\
Austin, USA\\
gbiros@acm.org} 
}

\maketitle

\thispagestyle{fancy}
\lhead{}
\rhead{}
\chead{}
\lfoot{\footnotesize{
SC24, November 17-22, 2024, Atlanta, Georgia, USA
\newline 979-8-3503-5291-7/24/\$31.00 \copyright 2024 IEEE}} 
\rfoot{}
\cfoot{}
\renewcommand{\headrulewidth}{0pt} 
\renewcommand{\footrulewidth}{0pt}

\begin{abstract}
  FIRAL is a recently proposed deterministic  active learning algorithm for multiclass classification using logistic regression. It was shown to outperform the state-of-the-art in terms of accuracy and robustness and comes with theoretical performance guarantees. However, its scalability suffers when dealing with datasets featuring a large number of points $n$, dimensions $d$, and classes $c$,  due to its $\mathcal{O}(c^2d^2+nc^2d)$ storage  and $\mathcal{O}(c^3(nd^2 + bd^3 + bn))$ computational complexity where $b$ is the number of points to select in active learning. To address these challenges, we propose an approximate algorithm with storage requirements reduced to $\mathcal{O}(n(d+c) + cd^2)$ and a computational complexity of $\mathcal{O}(bncd^2)$. Additionally, we present a parallel implementation on GPUs. We demonstrate the accuracy and scalability of our approach using MNIST,  CIFAR-10, Caltech101, and ImageNet. The accuracy tests reveal no deterioration in accuracy compared to FIRAL. We report strong and weak scaling tests on up to 12 GPUs, for three million point synthetic dataset.
\end{abstract}


\section{Introduction}\label{s:intro}

Let $\Xo$ be a set of labeled points  and $\Xu$ a set of $n$ unlabeled points, both sets sampled from the same distribution. We denote a labeled sample as a pair $(x,y)$, where $x\in\mathbb{R}^d$ is a point and $y \in \{1,2,\cdots, c \}$ is its label, where $c$ is the number of classes. Our goal of  \emph{active learning}  is to select $b$ points from $\Xu$ to label and use them along with pairs in $\Xo$ to train a multiclass logistic regression classifier.

Labeling data can be costly, but recent advancements in unsupervised and representation learning~\cite{bengio2013} enable us to leverage pre-existing feature embeddings combined with shallow learning techniques like logistic regression to develop efficient classification methods~\cite{simclr,zhuang2020}. The question is how to select training samples. Active learning addresses this issue by focusing on sample selection~\cite{ren2021survey}. Basic and popular sample selection methods include  random sampling and  k-means clustering. While these methods are scalable and easy to implement, they can be suboptimal and exhibit high variability due to their inherent randomness, particularly when the labeling budget is limited. We are seeking a method that is scalable, has low variability, and provides accuracy guarantees.

We propose a method for solving this problem based on the FIRAL algorithm (Fisher Information Ratio Active Learning) that appeared in 2023~\cite{firal-neurips}. FIRAL is an active learning algorithm with theoretical guarantees that outperforms the state of the art  in terms of accuracy. However, FIRAL has high complexity due to dense computations. Here we propose an approximate algorithm that dramatically accelerates FIRAL.  We dub the new algorithm \nfiral. A cornerstone in FIRAL is the Fisher information matrix, which is the \emph{Hessian  of a negative log-likelihood loss function}. In \nfiral we exploit the structure of the Hessian and we introduce the following: a matrix-free matrix-vector multiplication ``matvec'', a preconditioner, randomized trace estimators, and a modified regret minimization scheme. Overall the new components dramatically improve the complexity of the scheme. Combined with GPU and distributed memory parallelism \nfiral enables active learning for datasets that were intractable for FIRAL.
Our contributions can be summarized as follows:
\begin{itemize}
\item We  exploit structure, randomized linear algebra, and iterative methods to accelerate FIRAL. 
\item Using Python and CuPy~\cite{cupy17},and MPI~\cite{gropp-mpi99,dalcin2021mpi4py} we support multi-GPU acceleration. Our Python code is open-sourced.
\item We compare the accuracy of \nfiral with the exact FIRAL algorithm as well as several other popular active learning methods; and we test its scalability on multi-GPU systems. 
\end{itemize}
We further test the sensitivity of the method on different input parameters like the dataset size and the number of classes. Overall \nfiral is orders of magnitude faster that FIRAL without any noticeable difference in accuracy. While FIRAL is limited to datasets with a few thousands of points and up to 50 classes we demonstrate scalability to ImageNet 1.3 million points and 1000 classes, as well as synthetic datasets with  several million points. 

{\bf Related work:} There is a substantial body of work on active learning, including approaches such as uncertainty estimation~\cite{Li2013}, sample diversity~ \cite{Sener2017,Gissin2019discriminative,citovsky2021batch}, Bayesian inference~\cite{Gal2017,Pinsler2019bayesian}, and others. However, these methods lack performance guarantees. FIRAL provides lower and upper bounds of the generalization error for a multinomial logistic regression classifier assuming that the input points follow a sub-Gaussian distribution. It uses convex relaxation (\Relax step), similar to compressed sensing, to first compute weights for each point in $\Xu$ and then uses regret minimization to select  $b$ points (\Round step). Regarding parallel algorithms and GPU implementations, there are many implementations of random sampling and k-means and related combinations but nothing related to FIRAL-like algorithms.

{\bf Outline of the paper:} We start with the formulation of FIRAL in \cref{s:methods}. We summarize the \Relax step in \cref{s:ex-relax} and the \Round step  in \cref{s:ex-round}. The storage and computational complexity of FIRAL are summarized in~\cref{s:ex-firal-complexity}. We introduce \nfiral in  \cref{s:nfiral}: the Hessian structure and the accelerated \Relax step are described in \cref{s:ap-relax}; and the \Round solve is described in \cref{s:ap-round}. The HPC implementation and  complexity analysis are described in \cref{s:ap-hpc}. We report results from numerical experiments in~\cref{s:result}:  accuracy and comparisons with other active learning methods are reported in \cref{s:result-acc}; and  single and multi-GPU performance results are reported in \cref{s:result-single} and \cref{s:result-parallel} respectively.

\section{The exact FIRAL algorithm}\label{s:methods}

\subsection{Formulation} \label{s:formulation}
A summary of the main notation used in the paper can be found in \cref{table:notation}. We consider the batch active learning problem with given initial labeled points $\Xo$ and a pool of $n$ unlabeled points $\Xu$.  We denote a labeled sample as a pair $(x,y)$, where $x\in\mathbb{R}^d$ is a data point, $y \in \{1,2,\cdots, c \}$ is its label, and $c$ is the number of classes. We use a multiclass logistic regression model as our classifier.
 Given $x$ and classifier weights $\theta \in \mathbb{R}^{d\times (c-1)}$, the likelihood of a point $x$ having  label $y$ is defined by
\begin{align}\label{e:setup-conditional}
    p(y|x,\theta) = \begin{cases}
   \frac{\exp(\theta_y^\top x)}{1 + \sum_{l\in[c-1]} \exp(\theta_l^\top x)} ,\qquad y \in [c-1]\\
    \frac{1}{1 + \sum_{l\in[c-1]} \exp(\theta_l^\top x)},\qquad y = c.
    \end{cases}
\end{align}
We denote the vector of all class probabilities for point $x$ by $h(x) \in \mathbb{R}^{c-1}$, with $h_i = p(y=i | x)$.   To simplify notation we define $\td = d(c-1)$.
The weights $\theta$ are found by minimizing the negative log-likelihood: $\ell_{(x,y)}(\theta) \triangleq -\log p(y|x,\theta)$. The  Hessian or Fisher information matrix  at $x$ is defined by
 $\bH_i := \partial_{\theta \theta}\ell_{(x,y)}\in \mathbb{R}^{\td \times \td}$ and for our classifier is given by 
\begin{align}\label{eq:hessian}
    \bH_i = [\mathrm{diag}(h_i) - h_i h_i^\top] \otimes (x_i x_i^\top).
\end{align}

Let $\Ho$ be the summation of Hessians of the initial labeled points, $\Hp$ of the unlabeled points, and $\Hz$ of  \emph{weighted} unlabeled points with weights $z\in\mathbb{R}^n$, i.e.
\begin{align}\label{eq:sum-hessians}
    \Ho \triangleq \sum_{i\in \Xo} \Hi,\quad \Hp \triangleq  \sum_{i\in \Xu} \Hi, \quad \Hz \triangleq  \sum_{i \in \Xu} z_i \Hi.
\end{align}

Then given a budget of  $b$ points to sample (from $\Xu$), an optimal way would be to minimize the  Fisher Information Ratio~\cite{firal-neurips}: 
\begin{align}\label{eq:obj-all}
    \argmin_{z \in \{ 0,1\}^n, \|z\|_1= b } \big(\Ho + \Hz \big)^{-1} \cdot \Hp \triangleq f(z).
\end{align}
where ``$\cdot$" represents the matrix inner product. Unfortunately, this is an NP-hard combinatorial convex optimization problem. FIRAL proposed an algorithm to solve this problem with near-optimal performance guarantees. The algorithm is composed of two parts: a \Relax step of continuous convex relaxation optimization followed by a  \Round step to select  $b$ points. 

{
\begin{table}[!t]
\footnotesize
    \centering
    \caption{Summary of notation.}
	\label{table:notation}
    \begin{tabularx}{\columnwidth}{>{\hsize=0.2\hsize}X|
                              >{\hsize=0.7\hsize}X}
        \hline
        Notation     & Description     \\ \hline
         $d, c$ & dimension of point, number of classes \\ \hline
         $\td$ & $d \, c$ \\ \hline
        $b$ & budget: number of points to select for labeling \\ \hline
        $n$ &  number of points in unlabeled pool \\ \hline
        $\otimes$ & matrix Kronecker product  \\ \hline
        $\odot$ &  element-wise multiplication between two vectors \\\hline
        $vec(\cdot)$ &  vectorization of a matrix by stacking its columns  \\\hline
        $\Xo, \Xu$ & index sets for initial labeled points and unlabeled points \\ \hline
        $\Hi$ & Fisher information matrix for point $i$ (\cref{eq:hessian}) \\ \hline
        $\Ho, \Hp, \Hz$ & (weighted) sum of Hessians (\cref{eq:sum-hessians}) \\ \hline
        $\bSigma_z$ & sum of Hessians on selected points (\cref{eq:sigma}) \\ \hline
        $f(z)$ & objective function (\cref{eq:obj-all})  \\ \hline
        $g_i$&  gradient of relaxed objective  (\cref{eq:grad})\\ \hline
        $\mathcal{B}(\cdot)$ & block diagonal operation (\cref{def:block}) \\ \hline
        $z_\diamond$ & solution of relaxed problem (\cref{eq:obj-relax}) \\ \hline
        $\widetilde{\cdot}$ & matrix transformation (\cref{eq:mat-transform})\\ \hline
        $\eta$ & learning rate in round solver\\ \hline
        $\A_t$ & matrix in \Round step (\cref{eq:FTRL-action})\\ \hline
        $\B_t$  &   matrix used in \nfiral (\cref{eq:obj-round-sparse})\\\hline
    \end{tabularx}
\end{table}
}

\subsection{FIRAL:  \Relax step}\label{s:ex-relax}

The first step is to solve a continuous convex optimization problem which is formed by relaxing the constraint for $z$ in \cref{eq:obj-all}:
\begin{align}\label{eq:obj-relax}
   z_\diamond \in \argmin_{z \in [ 0,1]^n, \|z\|_1= b } \big(\Ho + \Hz \big)^{-1} \cdot \Hp.
\end{align}
The gradient of the objective w.r.t $z_i$ is 
\begin{align}\label{eq:grad}
    g_i =\frac{\partial f(z)}{\partial z_i} = - \Hi \cdot \bSigma_z^{-1} \Hp \bSigma_z^{-1},
\end{align}
where we define 
\begin{align}\label{eq:sigma}
    \bSigma_z = \Ho + \Hz.
\end{align}
FIRAL uses an entropic mirror descent algorithm to solve the relaxed problem.

\subsection{FIRAL: \Round step} \label{s:ex-round}
After the \cref{eq:obj-relax} step, FIRAL rounds  $z_\diamond$ into a valid solution to \cref{eq:obj-all} via regret minimization. Let us denote $\bSigma_\diamond = \Ho + \bH_{z_\diamond}$ and for any matrix $\bH \in \mathbb{R}^{\td \times \td}$, we define $\tH$ by
\begin{align}\label{eq:mat-transform}
    \tH \triangleq \bSigmahalfinv \bH \bSigmahalfinv.
\end{align}
The round solve has $b$ iterations and at each iteration $t\in[b]$, it selects the point $i_t$ s.t.
\begin{align}\label{eq:obj-round}
    i_t \in \argmin_{i \in \Xu} \Tr[(\A_t + \frac{\eta}{b} \tHo + \eta \tHi)^{-1}],
\end{align}
where $\eta>0$ is a hyperparameters (the learning rate), and  $\A_t \in\mathbb{R}^{\td \times \td}$ is a symmetric positive definite matrix defined by the Follow-The-Regularized-Leader algorithm:
\begin{align}\label{eq:FTRL-action}
    \A_t = \begin{cases}
    \sqrt{\td}\b I_{\td} & t=1\\
    \nu_t \b I  + \eta \tH_{t-1}& t>1
    \end{cases}, 
\end{align}
where $\nu_t\in\mathbb{R}$ is the unique constant s.t. $\Tr(\A_t^{-2}) = 1$, and 
\begin{align}
  \tH_{t-1} = \sum_{l=1}^{t-1}
  \left(\frac{1}{b} \tHo + \tH_{i_l}\right).
\end{align}

The FIRAL algorithm is near-optimal \cite{firal-neurips} in solving the optimization problem of \cref{eq:obj-all}:
\begin{theorem}\label{thm:firal}[Theorem 10 in \cite{firal-neurips}]
    Given $\epsilon \in (0,1)$, let $\eta = 8 \sqrt{\td}/\epsilon$, whenever $b\geq 32 \td/\epsilon^2 + 16 \sqrt{\td}/\epsilon^2$, denote $z$ as the solution corresponding to the points selected by \cref{algo:exact-firal}, then the algorithm is near-optimal: $f(z) \leq (1+\epsilon) f_*$, where $f_*$ is the optimal value of the $f$  in \cref{eq:obj-all}.
\end{theorem}

\subsection{Complexity and scalability of FIRAL}\label{s:ex-firal-complexity}
\cref{algo:exact-firal} summarizes FIRAL. Its storage complexity  is $\mathcal{O}(c^2d^2 + nc^2d)$ (\cref{table:complexity}),  which is prohibitively large for large $n$, $d$ or $c$. Furthermore, both relax and round solvers involve calculating inverse matrix of size $cd \times cd$. 
Thus, a scalable algorithm of FIRAL is needed.

{%
\begin{table*}[!t]
  \centering
  \footnotesize
    \caption{Comparison of algorithm complexity between FIRAL and \nfiral. $n_{\mathrm{relax}}$ is the number of mirror descent iterations in relax solver, $n_{\mathrm{CG}}$ is the number of CG iterations in each mirror descent step of the \nfiral relax solver.}
	\label{table:complexity}
  \begin{tabular}{@{}C{2.5cm}@{}|@{}C{2.9cm}@{}@{}C{2.8cm}@{}|C{3.5cm}@{}C{2.8cm}@{}}
    \toprule
   \multirow{2}{*}{Complexity} & \multicolumn{2}{c|}{Exact-FIRAL}  & \multicolumn{2}{c}{Approx-FIRAL}\\
    & Relax & Round  & Relax & Round \\
    \midrule
    Storage & $\mathcal{O}(c^2 d^2 + nc^2d)$ &$\mathcal{O}( c^2 d^2+ nc^2 d)$  & $\mathcal{O}(n(d+sc) + cd^2 )$ & $\mathcal{O}(n(d +c) +cd^2)$ \\ \hline
    Computation &$\mathcal{O}\big(n_{\mathrm{relax}}n c^3 d^2 \big)$  & $\mathcal{O}\big(b c^3(d^3  + n)\big)$  &$\mathcal{O}\big(n_{\mathrm{relax}} ncd(d + n_{\mathrm{CG}} s)\big)$  & $\mathcal{O}\big(b ncd^2\big)$ \\
    \bottomrule
  \end{tabular}
\end{table*}}

\begin{algorithm}[!t]
\footnotesize
\caption{\textproc{Exact-FIRAL}}
\label{algo:exact-firal}
\begin{algorithmic}[1]
    \State \centerline{\textcolor{gray}{\textit{\Relax step:}}}
    \State $z = (1/n, 1/n, \cdots, 1/n)\in \mathbb{R}^n$
    \State $\{\beta_t\}_{t=1}^T:$ schedule of learning rate for relax solve
\For {$t = 1$ to $T$}  \hfill \textcolor{lightgray}{\# $T$ is iteration number}
    \State $\bSigma_z \gets \Ho + \Hz $
    \State $ g_i \gets -\Tr(\Hi\bSigma_z^{-1} \Hp \bSigma_z^{-1}),\quad  \forall i \in[n]$ 
    \State $z_i \gets z_i \exp(-\beta_t g_i)$
    \State $z_i \gets \frac{z_i}{ \sum_{j\in [n]} z_j}$
\EndFor
\State $z_{\diamond} \gets b z$
\State \centerline{\textcolor{gray}{\textit{\Round step:}}}
\State $X\gets \emptyset$, $\bSigma_\diamond \gets \Ho + \bH_{z_\diamond}$
\State $\A_1\gets \sqrt{\td} \b I_{\td}$, $\tH\gets \b 0$
\For {$t= 1$ to $b$}
    \State $i_t\gets \argmin_{i \in \Xu} \Tr[(\A_t + \frac{\eta}{b} \Ho + \eta \Hi)^{-1}]$
    \State $\tH \gets \tH + \frac{1}{b}\tHo + \tH_{i_t}$
    \State $\b V \bLambda\b V^\top \gets $ eigendecomposition of $\eta \tH$
    \State find $\nu_{t+1}$ s.t. $\sum_{j \in[\td]} (\nu_{t+1} + \lambda_j)^{-2} = 1$ \textcolor{lightgray}{\# bisection}
    \State $\A_{t+1} \gets \b V (\nu_{t+1}\b I_{\td}  + \bLambda)\b V^\top $
    \State $X \gets X\cup \{ x_{i_t}\}$
\EndFor
\end{algorithmic}

\end{algorithm}


\section{The \nfiral algorithm} \label{s:nfiral}

\subsection{The Hessian structure and a fast \Relax step} \label{s:ap-relax}

The new \Relax solver has four components. First, we replace the exact trace operator in line 6 of \cref{algo:exact-firal} with a randomized trace estimator that only requires matvec operations. Second, we replace the direct solvers with a matrix-free conjugate gradients iterative method (CG). Third, we devise an exact fast matvec approximation for the Hessians. And fourth, we propose an effective preconditioner for the CG scheme. Taken together these components result in a scalable algorithm. We present the pseudo-code for our fast \Relax step in \cref{algo:new_relax} and summarize its complexity in \cref{table:complexity}.

We first develop an estimator for the gradient $g_i$ in \cref{eq:grad} that avoids constructing dense $\td$-by-$\td$ matrices such as $\bSigma_z$, $\Hp$, and $\bSigma_z^{-1}$.
The main idea is to use the Hutchinson trace estimator~\cite{Hutchinson1990} to approximate the gradient: suppose that we use $s$ Rademacher random vectors $\{ v_j\in\mathbb{R}^d\}_{j\in[s]}$, then $g_i$ can be approximated by
\begin{align}\label{eq:grad-est}
    g_i \approx -\frac{1}{s}\sum_{j \in [s]} v_j^\top \Hi (\bSigma_z^{-1}\Hp \bSigma_z^{-1} v_j).
\end{align}
To calculate the vector $\bSigma_z^{-1}\Hp \bSigma_z^{-1} v_j$ in \cref{eq:grad-est}, we can solve two linear systems using CG. Note that this term can be shared for all $i\in \Xu$ in gradient approximation formula. Thus, we only need to calculate the vector once for each  mirror descent iteration step.

\textbf{Fast matrix-free matvec.} The trace estimator and CG solvers require Hessian matvecs. The following Lemma gives an exact  closed form of the matvec  without forming the Hessian matrix explicitly.
\begin{lemma}[Matrix-free Hessian matvec]\label{lm:matvec}
For any given vector $v\in\mathbb{R}^{dc}$,  let $\b V\in\mathbb{R}^{d\times c}$ be the reshaped matrix from $v$ such that $vec(\b V) = v$. Denote the $j$-th column of $\b V$ by $v_j\in\mathbb{R}^d$, $k$-th component of $h_i$ by $h_i^k$. $\Hi$ is given by \cref{eq:hessian}. Then 
    \begin{align}
        \Hi v = \begin{bmatrix}
(x_i^\top v_1 - x_i^\top\b V  h_i) h_{i}^1 x_i \in \mathbb{R}^d\\
\vdots\\
(x_i^\top v_c - x_i^\top \b V h_i) h_{i}^c x_i \in \mathbb{R}^d
\end{bmatrix} \in \mathbb{R}^{\td}.\nonumber
    \end{align}
\end{lemma}
\begin{proof}
   \begin{align}
       \Hi v &= [\mathrm{diag}(h_i) \otimes (x_i x_i^\top)] v - [(h_i h_i^\top) \otimes (x_i x_i^\top)] v \nonumber\\
       &= vec\big(x_i x_i^\top \b V \mathrm{diag}(h_i) \big) - vec\big(x_i x_i^\top \b V h_i h_i^\top \big) \nonumber\\
       & = vec\big( \big[(x_i^\top v_1) h_{i}^1 x_i, \cdots , (x_i^\top v_c) h_{i}^c x_i\big]\big) \nonumber\\
       & - (x_i^\top \b V h_i) vec(x_i h_i^\top),\nonumber
   \end{align} 
   where the second equality uses a property of the matrix Kronecker product.
\end{proof}

According to \cref{lm:matvec}, we can compute $\Hi v$ in the following steps: \nbone $\gamma_i \gets \b V^\top x_i$, \nbtwo $\alpha_i \gets \gamma_i^\top h_i$, \nbthree $\gamma_i \gets (\gamma_i - \alpha_i) \odot h_i$, and \nbfour $\Hi v \gets vec(\gamma_i \otimes x_i)$. It is worth noting that the storage required for the first three steps is only $c+1$ elements, while the last step requires $dc$ elements  for storing the result of the matvec operation. A comparison of the complexity between our fast matvec algorithm and direct matvec is provided in \cref{table:compare-matvec}.

\begin{table}[!t]
\footnotesize
    \centering
    \setlength{\tabcolsep}{3pt}
 \caption{Comparison of storage and computational complexity between matrix-free matvec and direct matvec.}
   \label{table:compare-matvec}
    \begin{tabular}{c|c|c}
      \toprule
     method  &storage &computation\\\midrule
        direct MatVec &$\mathcal{O}(d^2c^2)$ & $\mathcal{O}(d^2c^2)$\\ 
         fast MatVec & $\mathcal{O}(dc)$ &$ \mathcal{O}(dc) $ \\ 
        \bottomrule
    \end{tabular}
    \label{tab:my_label}
\end{table}

With the help of the matrix-free matvec, we can calculate $\Hp v$ by 
\begin{align}
    \Hp v = \begin{bmatrix}
\sum_{i \in \Xu}\gamma_i^1 x_i \\
\vdots\\
\sum_{i \in \Xu}\gamma_i^c  x_i ,
\end{bmatrix} 
\end{align}
where $\gamma_i^k = (x_i^\top v_1 - x_i^\top\b V  h_i) h_{i}^k$ for $k\in [c]$. Based on the previous analysis, the additional storage required is solely for $\gamma_i$ for all unlabeled points $\Xu$, amounting to $4n(c+1)$ memory cost. We can use the similar calculation for the matvec of $\bSigma_z v$ within the CG iterations.

\textbf{Preconditioned CG.} To further  accelerate the calculation, we propose a simple but, as we will see, effective block diagonal preconditioner for the CG solves. We first introduce the block diagonal operation as follows.
\begin{definition}[Block diagonal operation $\mathcal{B}(\cdot)$]\label{def:block}
    For any matrix $\bH \in \mathbb{R}^{\td \times \td}$, define  $\mathcal{B}(\bH) \in \mathbb{R}^{\td \times \td}$ as the matrix comprising $d\times d$ block diagonals of $\bH$; denote the $k$-th block diagonal matrix by $\mathcal{B}_k(\bH) \in\mathbb{R}^{d\times d}$. 
\end{definition}
Then, for every Hessian matrix $\Hi$ in \cref{eq:hessian}, we have its block diagonal as
\begin{align}\label{eq:Hessian-block}
   \mathcal{B}( \Hi)= [\text{diag}(h_i \odot (1-h_i))] \otimes (x_i x_i^\top),
\end{align}
and its $k$-th matrix diagonal as
\begin{align}
    \mathcal{B}_k( \Hi) = h_i^k (1-h_i^k) \cdot x_i x_i^\top.
\end{align}

\begin{figure}[!t]
\centering
\begin{tikzpicture}
        \node[inner sep=0pt] (a) at (0,0) {\includegraphics[width=4cm]{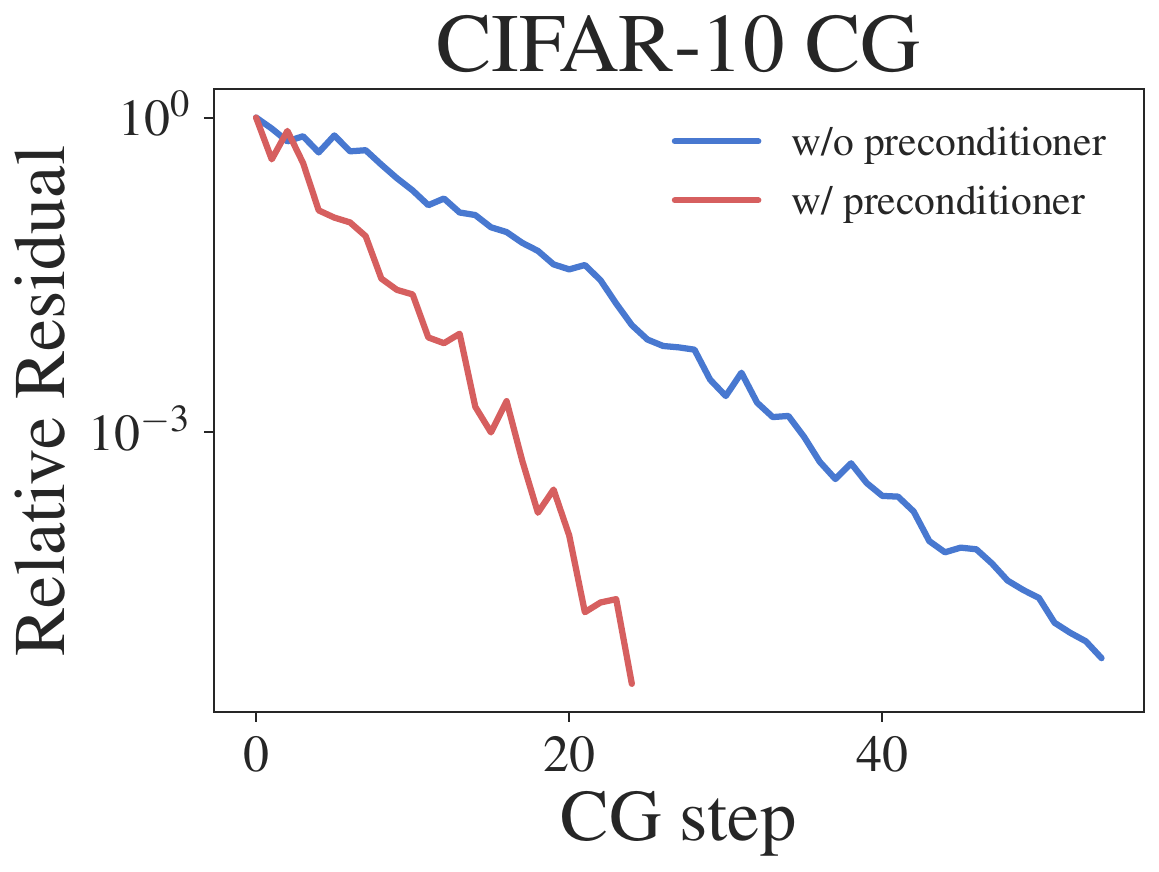}};
         \node[inner sep=0pt] (b) at (4.2,0) {\includegraphics[width=4cm]{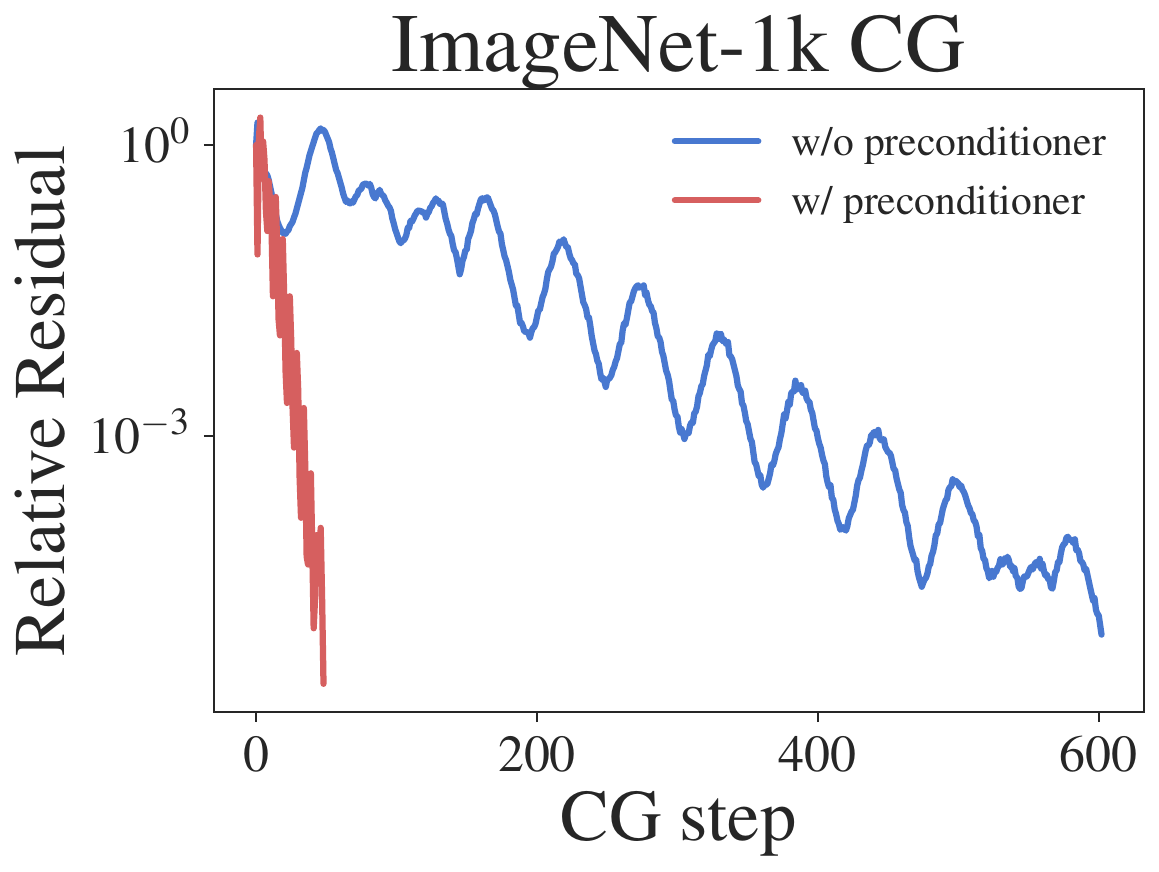}};  
\end{tikzpicture}
\caption{The impact of preconditioner on CG iterations. The experimental setup is detailed in \cref{s:result-acc}. We showcase the convergence of CG in the initial mirror descent iteration (i.e., Line 6 of \cref{algo:new_relax}).}
\label{fig:cg}
\end{figure}

We employ $\mathcal{B}(\bSigma_z)^{-1}$ as the preconditioner for CG to solve the linear system required  for gradient estimation in \cref{eq:grad-est}. We illustrate the effectiveness of the CG preconditioner for two datasets in \cref{fig:cg}. Using  $\mathcal{B}(\bSigma_z)^{-1}$ as preconditioner accelerates CG convergence due to several factors. Firstly, it improves the conditioning of the matrix. For instance, in the CIFAR-10 test, the condition number of $\bSigma_z$ is 198, while the condition number of $\mathcal{B}(\bSigma_z)^{-1} \bSigma_z$ is 72. Additionally, the majority of eigenvalues of the preconditioned matrix are clustered into small intervals.

\begin{algorithm}[!t]
\footnotesize
\caption{\textproc{Fast \Relax Solve}}
\label{algo:new_relax}
\begin{algorithmic}[1]
    \State $z = (1/n, 1/n, \cdots, 1/n)\in \mathbb{R}^n$
    \State $\{\beta_t\}_{t=1}^T:$ schedule of learning rate for relax solve
\For {$t = 1$ to $T$}  \hfill \textcolor{lightgray}{\# $T$ is iteration number}
\State $\b V = [v_1, v_2, \cdots, v_s]\in \mathbb{R}^{dc \times s}$: matrix of $s$ Rademacher random vectors.
    \State $\{\mathcal{B}_k(\bSigma_z)^{-1}\}_{k\in[c]}\gets$preconditioner for CG solve
    \State $\b W \gets \bSigma_z^{-1} \b V$ by preconditioned CG
    \State  $\b W \gets \Hp \b W$
    \State $\b W  \gets \bSigma_z^{-1} \b W$ by preconditioned CG
    \State $g_i \gets -\frac{1}{s}\sum_{j\in[s]} v_j^\top \Hi w_j, \qquad \forall i\in \Xu$
    \State $z_i \gets z_i \exp(-\beta_t g_i)$
    \State $z_i \gets \frac{z_i}{ \sum_{j\in [n]} z_j}$
\EndFor
\State $z_{\diamond} \gets b z$
\end{algorithmic}

\end{algorithm}

\subsection{The new \Round step} \label{s:ap-round}

\begin{algorithm}[!t]
\footnotesize
\caption{\textproc{Approx-FIRAL}}
\label{algo:approx-firal}
\begin{algorithmic}[1]
\State $z_{\diamond} \gets$ solution of \Relax step from \cref{algo:new_relax}
\State \centerline{\textcolor{gray}{\textit{Diagonal \Round step:}}}
\State $X\gets \emptyset$, form $\{ (\bSigma_\diamond)_k\in \mathbb{R}^{d\times d}\}_{k\in[c]}$
\State $ \big\{ (\B_1)_k^{-1}\gets [\sqrt{\td} (\bSigma_\diamond)_k + \frac{\eta}{b} (\Ho)_k]^{-1}\big\}_{k\in[c]} $
\State $\big\{ (\bH)_k\gets \b 0\big \}_{k\in[c]}$
\For {$t= 1$ to $b$}
    \State $i_t\gets$ \cref{eq:obj-round-sparse}
    \State $\big \{ (\bH)_k \gets (\bH)_k + \frac{1}{b}(\Ho)_k + h_{i_t}^k (1- h_{i_t}^k) x_{i_t} x_{i_t}^\top \big \}_{k \in[c]}$
    \State $\big\{ [\lambda_{k,j}]_{j=1}^d \gets \text{ eigenvalues of }  (\tH)_k\big\}_{k\in[c]}$
    \State find $\nu_{t+1}$ s.t. $\sum_{k\in[c]}\sum_{j \in[d]} (\nu_{t+1} +\eta \lambda_{k,j})^{-2} = 1$ 
    \State $\big\{ (\B_{t+1})_k^{-1} \gets [\nu_{t+1}(\bSigma_\diamond)_k + \eta (\bH)_k +  \frac{\eta}{b}(\Ho)_k ]^{-1} \big\}_{k\in[c]}$
    \State $X \gets X\cup \{ x_{i_t}\}$
\EndFor
\end{algorithmic}

\end{algorithm}

The difficulty of the \Round step lies in computing the objective value in \cref{eq:obj-round} for each point $i\in\Xu$ at each round $t\in[b]$. Even when employing CG with the fast matrix-free matvec introduced in the preceding section, the computational complexity for estimating the objective is $\mathcal{O}(bn_{\mathrm{CG}} n^2 cds)$, which is prohibitively large for large-scale problems.

Motivated by the effectiveness of the preconditioner in the \Relax, it is natural to consider some approximation. Notice that the \Round step becomes much easier when considering only the block diagonals of all Hessian matrices. Specifically, we assume that each Hessian matrix $\bH_i$ retains only its block diagonal parts, as expressed in \cref{eq:Hessian-block}. Consequently, all matrices with a size of $\td\times \td$ in the \Round step are block diagonal. This assumption not only reduces storage requirements but also simplifies calculations. Firstly, we introduce a Sherman-Morrison-like formula for the low-rank updates for the inverse of a block diagonal matrix in \cref{lm:sherman-morrison}. Subsequently, we present a simple yet equivalent objective to the original exact \Round step in \cref{prop:sparse-round}. We outline the pseudo-code in \cref{algo:approx-firal} and summarize the complexity of the new \Round step in \cref{table:complexity}.

\begin{lemma}\label{lm:sherman-morrison}
Let $\A\in\mathbb{R}^{\td \times \td}$ be a block diagonal positive definite matrix with $c$ block diagonals of $d \times d$ matrices, $x \in\mathbb{R}^d$ and $\gamma \in \mathbb{R}^c$ be vectors. If $\A + \mathrm{diag}(\gamma) \otimes (x x^\top)$ is positive definite, then $\big(\A + \mathrm{diag}(\gamma) \otimes (x x^\top)\big)^{-1}$ is a block diagonal matrix with its $k$-th block having the following form:
\begin{align}
    \big(\A + \mathrm{diag}(\gamma) \otimes (x x^\top) \big)_k^{-1} = \A_k^{-1} - \frac{\gamma_k \A_k^{-1}x x^\top \A_k^{-1}}{1 + \gamma_k x^\top \A_k^{-1} x},
\end{align}
where $\A_k^{-1}$ is the inverse of $k$-th diagonal of $\A$, $\gamma_k$ is the $k$-th component of $\gamma$.
\end{lemma}

\begin{proposition}\label{prop:sparse-round}
If all Fisher information matrices  $\Hi$ only preserve the block diagonals of $d\times d$ matrices, then at each iteration of the \Round step, the objective defined in \cref{eq:obj-round} is equivalent to the following:
\begin{align}\label{eq:obj-round-sparse}
         i_t \in \argmax_{i\in \Xu} \sum_{k=1}^c h_i^k (1-h_i^k)\cdot \frac{x_i^\top (\B_t)_k^{-1}(\bSigma_\diamond)_k^{-1} (\B_t)_k^{-1}x_i}{1 + \eta h_i^k (1-h_i^k) x_i^\top (\B_t)_k^{-1} x_i},
\end{align}
where $\B_t = \bSigma_\diamond^{1/2} \A_t \bSigma_\diamond^{1/2} + \frac{\eta}{b}\Ho$.
\end{proposition}

\begin{proof}
We denote the objective for point $i\in \Xu$ in round problem \cref{eq:obj-round} by $r_i$, then
\begin{align}
    r_i &= \Tr[(\A_t + \frac{\eta}{b} \tHo + \eta \tHi)^{-1}] \nonumber\\
        &= \Tr\Big[\bSigma_\diamond^{1/2} \big( \underbrace{\bSigma_\diamond^{1/2} \A_t
 \bSigma_\diamond^{1/2} + \frac{\eta}{b} \Ho}_{\triangleq \B_t} + \eta \Hi \big)^{-1} \bSigma_\diamond^{1/2}\Big]\nonumber\\
 & = \Tr\Big[\big(\B_t + \eta \Hi \big)^{-1}  \bSigma_\diamond\Big]. \label{eq:in-obj-round-1}
\end{align}
Since $\B_t$ and $\Hi$ are both block diagonal, by \cref{lm:sherman-morrison}, $k$-th block diagonal of  $\big(\B_t + \eta \Hi \big)^{-1}$ has the following form:
\begin{align}\label{eq:in-obj-round-2}
    \big(\B_t + \eta \Hi \big)_k^{-1} = \big(\B_t\big)_k^{-1} - \frac{\eta h_i^k (1-h_i^k) (\B_t)_k^{-1} x_i x_i^\top (\B_t)_k^{-1}  }{1 + \eta h_i^k (1-h_i^k) x_i^\top (\B_t)_k^{-1} x_i }.
\end{align}
Substitute \cref{eq:in-obj-round-2} into \cref{eq:in-obj-round-1}, we have 
\begin{align}
    r_i = &\Tr[\B_t^{-1} \bSigma_\diamond] \nonumber\\
    &- \eta \sum_{k=1}^c h_i^k (1-h_i^k)\cdot \frac{x_i^\top (\B_t)_k^{-1}(\bSigma_\diamond)_k^{-1} (\B_t)_k^{-1}x_i}{1 + \eta h_i^k (1-h_i^k) x_i^\top (\B_t)_k^{-1} x_i},
\end{align}
which leads to \cref{eq:obj-round-sparse}.
\end{proof}

\subsection{HPC implementation and complexity analysis}\label{s:ap-hpc}


{%
\begin{table*}[!t]
  \centering
  \footnotesize
    \caption{Storage, computation and communication complexity of parallel implementation of Approx-FIRAL (\cref{algo:approx-firal}). The detailed derivations are presented in \cref{s:ap-hpc}. $n_{\mathrm{relax}}$ represents the number of mirror descent iteration in \cref{algo:new_relax}, $n_{\mathrm{CG}}$ represents the number of CG iterations. }
	\label{table:parallel-complexity}
  \begin{tabular}{@{}C{2.5cm}@{}|@{}C{3.5cm}@{}|@{}C{4.2cm}@{}|@{}C{6.8cm}@{}}
    \toprule
 {Complexity} & Storage & Computation & Communication\\
 \midrule
   \Relax step &$\mathcal{O}\Big(\frac{n}{p}(d+c) + cds + cd^2\Big)$  &$\mathcal{O}\Big(n_{\mathrm{relax}} cd \big(\frac{n}{p} (d + n_{\mathrm{CG}}s) + d^2\big)\Big)$   & $\mathcal{O}\Big(n_{\mathrm{relax}}\log p \big( n_{\mathrm{CG}}  t_s + cd(n_{\mathrm{CG}} s + d) ( t_w + t_c) \big)\Big)$ \\ \hline
\Round step &$\mathcal{O}\Big(\frac{n}{p}(d+c) + cd^2\Big)$   & $\mathcal{O}\Big( b cd^2 (\frac{n}{p} +d) \Big)$ & $\mathcal{O}\Big( b \log p \big(t_s + (d+c) t_w + t_c \big)\Big)$ \\
    \bottomrule
  \end{tabular}
\end{table*}}

Our HPC implementation of \nfiral, as outlined in \cref{algo:new_relax,algo:approx-firal}, is GPU-based. We employ \texttt{cupy} \cite{cupy17} for computation and \texttt{mpi4py} \cite{dalcin2021mpi4py} for communication within GPUs. To utilize a GPU-aware Message Passing Interface (MPI), we utilize MVAPICH2-GDR \cite{mvapich}. Our implementation employs single-precision floating point for both storage and computation. Let $p$ be the number of GPUs, we start the parallel implementation by evenly distributing $h_i$ and $x_i$ of $n$ points in $\Xu$ across $p$ GPUs.

Regarding computation, we  utilize the built-in functions of the linear algebra routines available in \texttt{cupy}. We provide a summary of some of the key functions as follows:
\begin{itemize}
    \item \texttt{cupy.einsum}: In the \Relax step outlined in \cref{algo:new_relax}, we utilize Einstein summation to construct the block diagonal matrix as a preconditioner in Line 5. In Lines 6-8, we employ Einstein summation for the fast matrix-free matvec developed in \cref{s:ap-relax} for matrices $\bSigma_z$ and $\Hp$. For the \Round step in \cref{algo:approx-firal}, we use this function mainly for the objective calculation in \cref{eq:obj-round-sparse} (Line 7).
    \item \texttt{cupy.linalg.eigvalsh}: In the \Round step, this function is employed to compute the eigenvalues of the block diagonals of $\tH$ in a batch-wise manner in Line 9 of \cref{algo:approx-firal}. In our implementation, we evenly distribute the computation of eigenvalues for $c$ block diagonals among $p$ GPUs.
    \item \texttt{cupy.linalg.inv}: This function is utilized to calculate the inverse of block diagonal matrices in Line 5 of \cref{algo:new_relax} and Lines 4 and 11 of \cref{algo:approx-firal}.
\end{itemize}

As for communication among GPUs, we outline the primary collective communication operations utilized as follows:
\begin{itemize}
\item \texttt{MPI\_Allreduce}: For \Relax step in \cref{algo:new_relax}, we need this operation for summation of the block diagonals in Line 5. In Lines 6-8, it is necessary for the summation of the results from the matvec operation. For \Round step in \cref{algo:approx-firal}, we use \texttt{MPI\_Allreduce} in Line 7 to find the point with the global maximum objective value across all GPUs.
\item \texttt{MPI\_Allgather}: This operation is employed to collect all eigenvalues in the \Round step (Line 9 of \cref{algo:approx-firal}).
\item \texttt{MPI\_Bcast}: In Lines 6-8 of \cref{algo:new_relax}, we distribute $\b W$ to each GPU. In Line 11 of \cref{algo:approx-firal}, we utilize this operation to transmit $h_{i_t}$ and $x_{i_t}$ to all GPUs.
\end{itemize}

In \cref{table:parallel-complexity}, we summarize the complexity of storage, computation and communication for our HPC implementation of  \nfiral. The details are outlined as follows. 
To estimate the cost of collective communications, we rely on the results presented in \cite{Rajeev2005}. We assume that the time used to send a message between two processes is $t_s + m t_w$, where $t_s$ is the latency, $t_w$ is the transfer time per byte, and $m$ denotes the number of bytes transferred. Additionally, we denote the computation cost per byte by $t_c$ for performing the reduction operation locally on any process. The costs associated with the three MPI operations we utilized are as follows: \nbone \texttt{MPI\_Allreduce}: employing the recursive doubling algorithm, the time complexity is $\log p (t_s + m (t_w + t_c))$. \nbtwo \texttt{MPI\_Allgather}: utilizing the recursive doubling algorithm, the time complexity is $\log p t_s + \frac{p-1}{p} m t_w$. \nbthree \texttt{MPI\_Bcast}: using the binomial tree algorithm, the time complexity is $\log p (t_s + m t_w)$. \par\medskip

\textbf{\Relax step.}
In terms of storage, the parallel implementation of \cref{algo:new_relax} requires storing Rademacher random vectors $\b V$ (Line 4), the intermediate matrix $\b W$ (Lines 6-8), and the inverses of $c$ block-diagonal matrices (Line 5). Hence, the total storage for each GPU amounts to $\mathcal{O}\left(\frac{n}{p}(d+c) + cds + cd^2\right)$ including the storage of $x_i$ and $h_i$ for $\frac{n}{p}$ points. 

For building the preconditioner of CG (Line 5), each GPU initially computes the block diagonal matrices $\{\mathcal{B}(\bSigma_z)_k\}_{k\in[c]}$ with a complexity of $\mathcal{O}(\frac{n}{p} cd^2)$. The \texttt{MPI\_Allreduce} operation for aggregation of these matrices across all GPUs incur  a communication cost of $\mathcal{O}\left(\log p( t_s +cd^2 (t_w+  t_c)\right)$. Then each GPU calculates the inverse of the block diagonal matrices as the preconditioner, which has a computational complexity of $\mathcal{O}(cd^3)$. In summary, the computational and communication time required to construct the preconditioner are as follows:
\begin{align}
    T_{\mathcal{B}({\bSigma_z})}^{\mathrm{comp}}& = \mathcal{O}\big(cd^2\big(\frac{n}{p} + d\big)\big), \\
    T_{\mathcal{B}({\bSigma_z})}^{\mathrm{comm}} &= \mathcal{O}\big(\log p( t_s +cd^2 (t_w+  t_c)\big).
\end{align}

Within each preconditioned CG iteration (Lines 6 and 8), the primary time consumption arises from the matvec calculations of $\bSigma_z \b V$ and $\mathcal{B}(\bSigma_z)\b V$. According to the complexity outlined in our fast matvec algorithm in \cref{table:compare-matvec}, computation of matvec has a complexity of $\mathcal{O}\left(\frac{n}{p}cds\right)$. Subsequently, the summation of these vectors requires an \texttt{MPI\_Allreduce} operation with a communication cost of $\mathcal{O}\left(\log p (t_s +  cds (t_w + t_c)\right)$. The computation of $\mathcal{B}(\bSigma_z)\b V$ solely demands a computational cost of $\mathcal{O}(cd^2s)$. Let $n_{\mathrm{CG}}$ be the CG iteration number, we have 
\begin{align}
    T_{\mathrm{CG}}^{\mathrm{comp}}& = \mathcal{O}\Big(n_{\mathrm{CG}}\frac{n}{p} cds\Big), \\
    T_{\mathrm{CG}}^{\mathrm{comm}} &= \mathcal{O}\Big( n_{\mathrm{CG}}\log p (t_s +  cds (t_w+t_c))\Big).
\end{align}

Regarding other components of the relax solver, Line 7 of the matvec operation has a complexity similar to one step of CG. The computation of the gradient $g_i$ (Line 9) and the updating of $z$ (Lines 10-11) necessitate a complexity of $\mathcal{O}\left(\frac{n}{p} cds\right)$.\par\medskip

\textbf{\Round step.} Regarding storage, all matrices utilized in \cref{algo:approx-firal} are block diagonal matrices, resulting in a storage requirement of $\mathcal{O}(cd^2)$. Furthermore, to compute the objective for each point in Line 7, additional storage of $\mathcal{O}(nc)$ is necessary. As a result, the total storage requirement is $\mathcal{O}(n(c+d) + cd^2)$.

During each iteration of the \Round step, computing the objective function for each point in Line 7 (\cref{eq:obj-round-sparse}) needs a computational complexity of $\mathcal{O}(\frac{n}{p} cd^2)$. Subsequently, to select the point with the maximum objective, we utilize \texttt{MPI\_Allreduce} to gather and compare the maximum objective across local processes, resulting in a communication cost of $\mathcal{O}(\log p (t_s + t_w + t_c))$. 

To update $\{ (\b H)_k\}_{k\in[c]}$ (Line 8), the process owning $i_t$ broadcasts $x_{i_t}$ and $h_{i_t}$ to other processes using an \texttt{MPI\_Bcast} operation with a size of $\mathcal{O}(c+d)$.  In Line 9, we first compute eigenvalues for $\frac{c}{p}$ matrices for each process, followed by collecting all eigenvalues using \texttt{MPI\_Allgather}. The computational complexity of this step is $\mathcal{O}(\frac{c}{p}d^3)$, and the communication cost is $\mathcal{O}(\log p t_s + \frac{c}{p} t_w)$. As for Line 11, computing the inverse matrices requires a computational complexity of $\mathcal{O}(cd^3)$. The total computation and communication complexity for the \Round step are summarized in \cref{table:parallel-complexity}.

\section{Numerical Experiments}\label{s:result}
We test the classification accuracy in  \cref{s:result-acc}, single node performance in \cref{s:result-single} and parallel computing performance in \cref{s:result-parallel} on the Lonestar6 A100 nodes in the Texas Advanced Computing Center (TACC). Lonestar6 A100 nodes are interconnected with IB HDR (200 Gbps) and have three A100 NVIDIA GPUs per node.

\subsection{Active learning performance}\label{s:result-acc}
\begin{table*}[!t]
\footnotesize
    \caption{Summary of datasets for active learning experiments. }
    \label{table:acc-data}
    \centering
    \begin{tabular}{c|c|c|c|c|c|c|c|c}
    \toprule
        Name & Type& \# classes &dimension & $\Xo$   & $\Xu$   &\# rounds   & budget/round   & \# evaluation points\\
    \midrule
MNIST & balanced & 10 &20  &10   &3,000   & 3   &  10  &60,000 \\\hline
CIFAR-10  & balanced & \multirow{2}{*}{10} &\multirow{2}{*}{20}  &\multirow{2}{*}{10}   &\multirow{2}{*}{3,000}  & \multirow{2}{*}{3} &  \multirow{2}{*}{10}  &\multirow{2}{*}{50,000}\\
imb-CIFAR-10 & imbalanced&  &  &  &  &  &   &\\\hline
ImageNet-50 & balanced& \multirow{2}{*}{50} &\multirow{2}{*}{50}  &\multirow{2}{*}{50}   &\multirow{2}{*}{5,000}  & \multirow{2}{*}{6} &  \multirow{2}{*}{50}  &\multirow{2}{*}{64,273}\\
imb-ImageNet-50 & imbalanced&  &  &  &  &  &   &\\\hline
Caltech-101  & imbalanced& 101 &100  &101   &1,715   & 6   &  101  &8,677 \\\hline
ImageNet-1k  & balanced & 1,000 &383  &2,000   &50,000   & 5  &  200  &1,281,167 \\
    \bottomrule
    \end{tabular}
\end{table*}

\begin{table}[!t]
\footnotesize
    \caption{Time comparison between \exactfiral and \nfiral on a single A100 GPU. The time reported in the table is in seconds.}
    \label{table:acc-time}
    \centering
\begin{tabular}{ccc}
    \toprule
     & \exactfiral &\nfiral \\\midrule
 \rowcolor{LightCyan}     \multicolumn{3}{c}{ImageNet-50} \\
     \Relax &33.6   &1.3 \\
     \Round &34.8   &1.1 \\\hline
 \rowcolor{LightCyan}    \multicolumn{3}{c}{Caltech-101} \\
     \Relax & 172.3 & 1.9\\
     \Round & 945.3 & 4.4\\
     \bottomrule
\end{tabular}
\end{table}

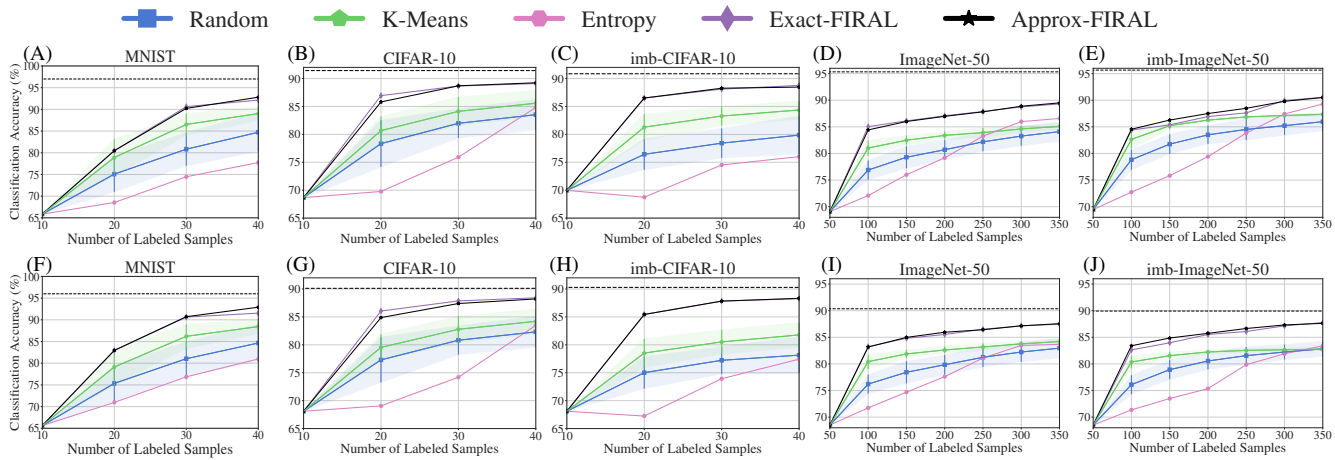
\begin{figure*}
\centering
\begin{tikzpicture}
\node[inner sep=0pt] (a1) at (3.,1.7) {\includegraphics[width=14cm]{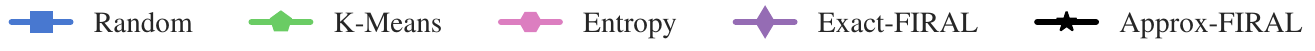}};
    \node[inner sep=0pt] (a0) at (-3.7,0) {\includegraphics[width=3.5cm,height=2.75cm]{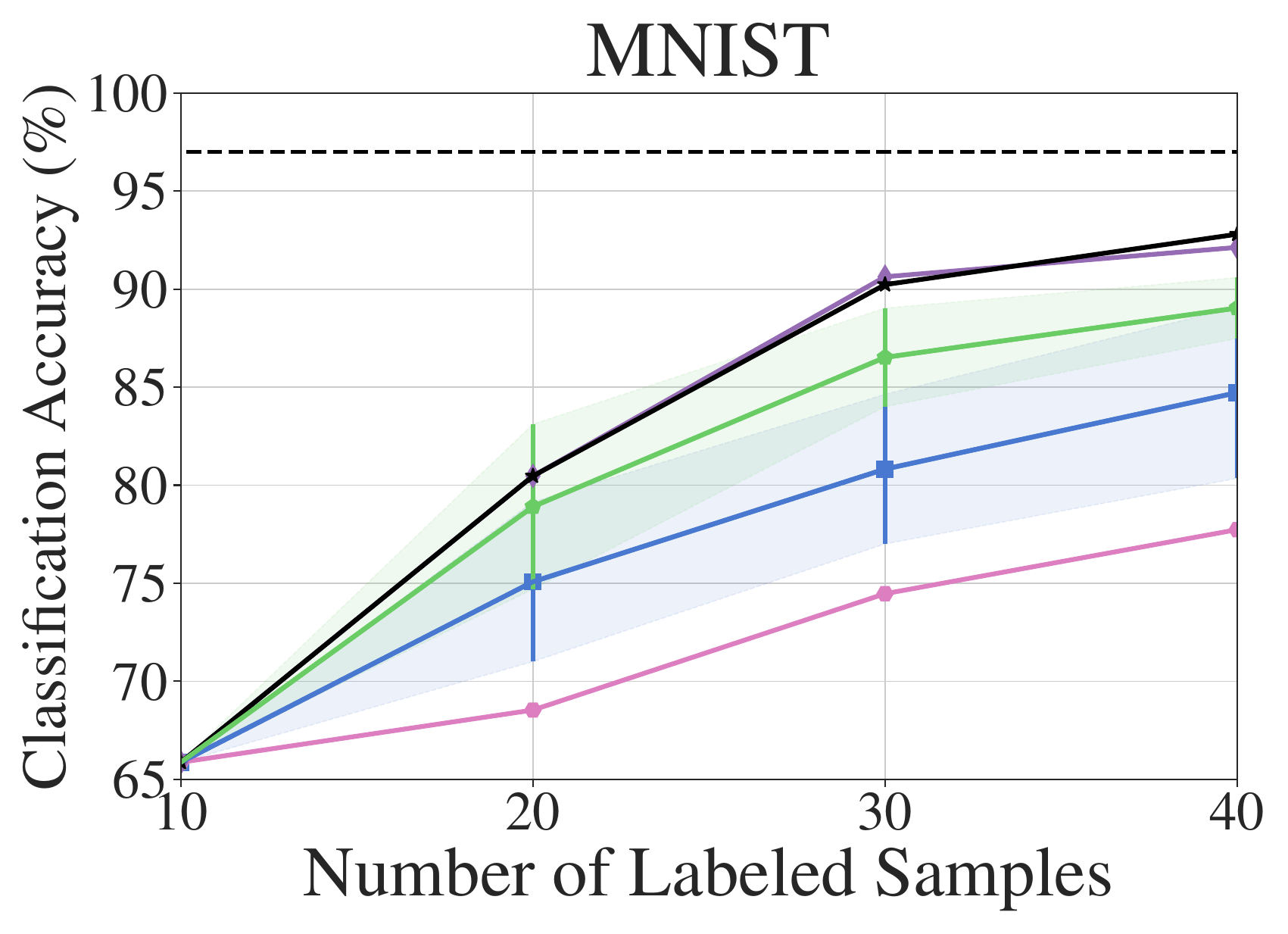}};
    \node[inner sep=0pt] (b0) at (-3.7,-2.8) {\includegraphics[width=3.5cm,height=2.75cm]{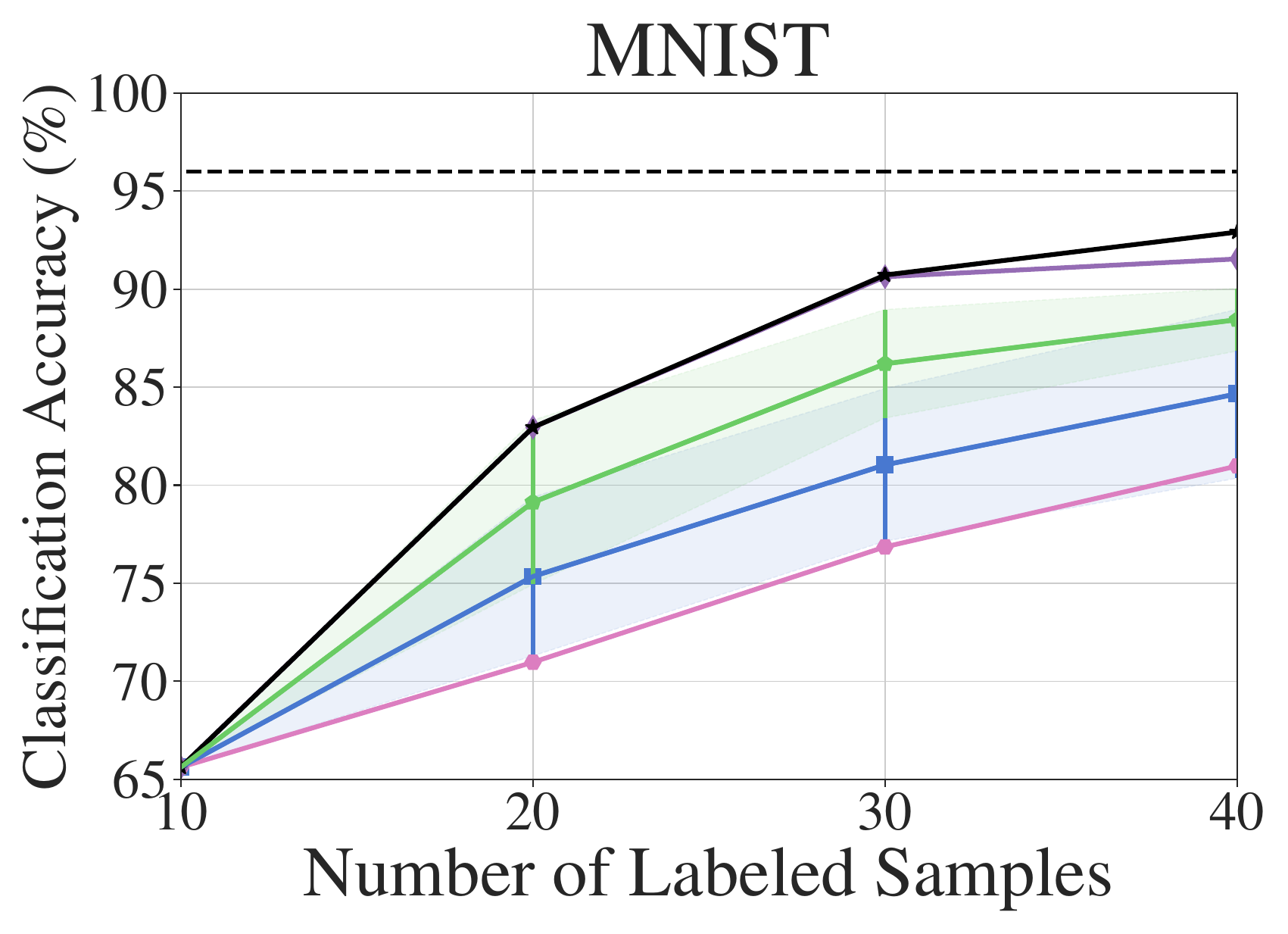}};
    \node[inner sep=0pt] (a1) at (0,0) {\includegraphics[width=3.5cm]{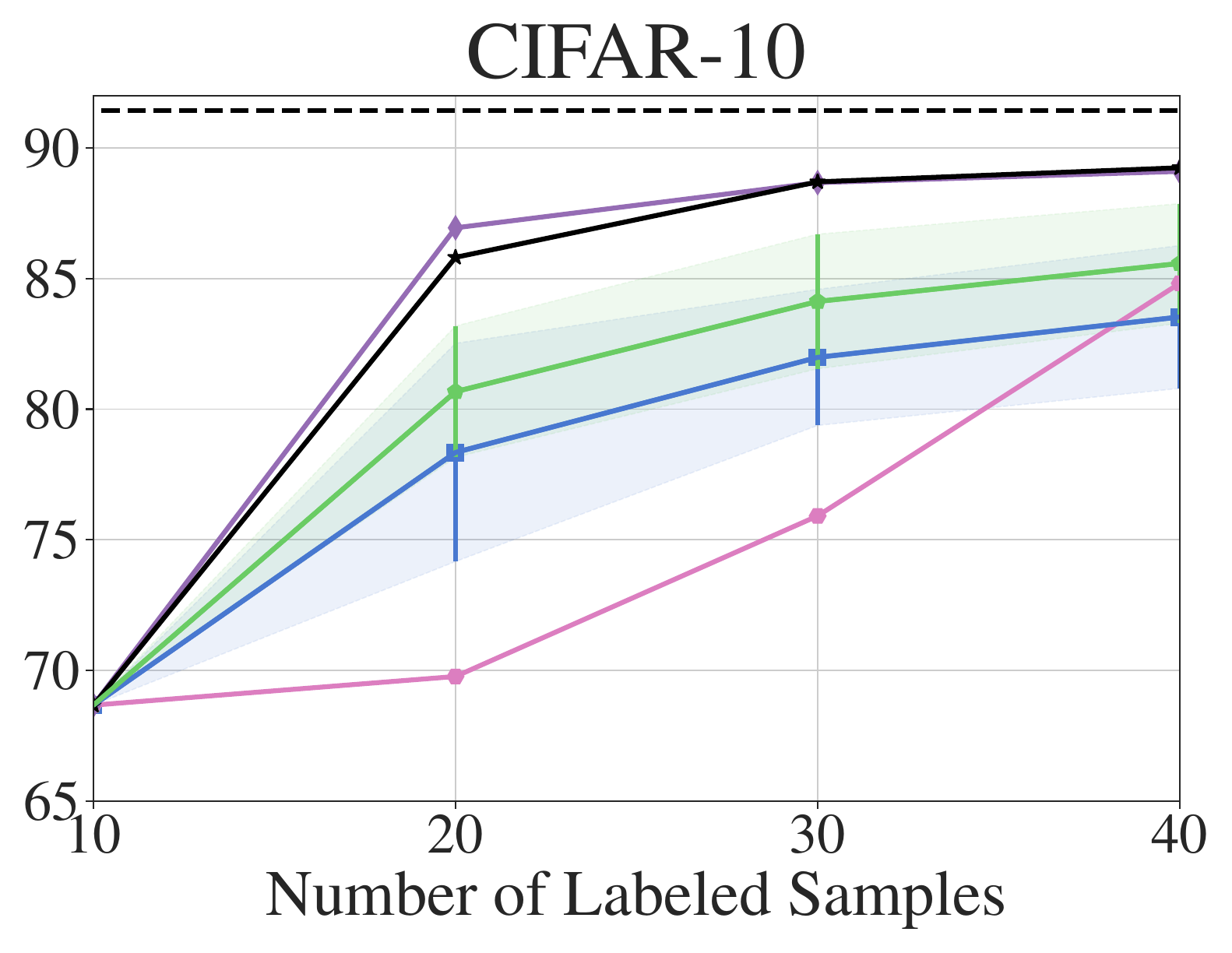}};
    \node[inner sep=0pt] (a2) at (3.5,0) {\includegraphics[width=3.5cm]{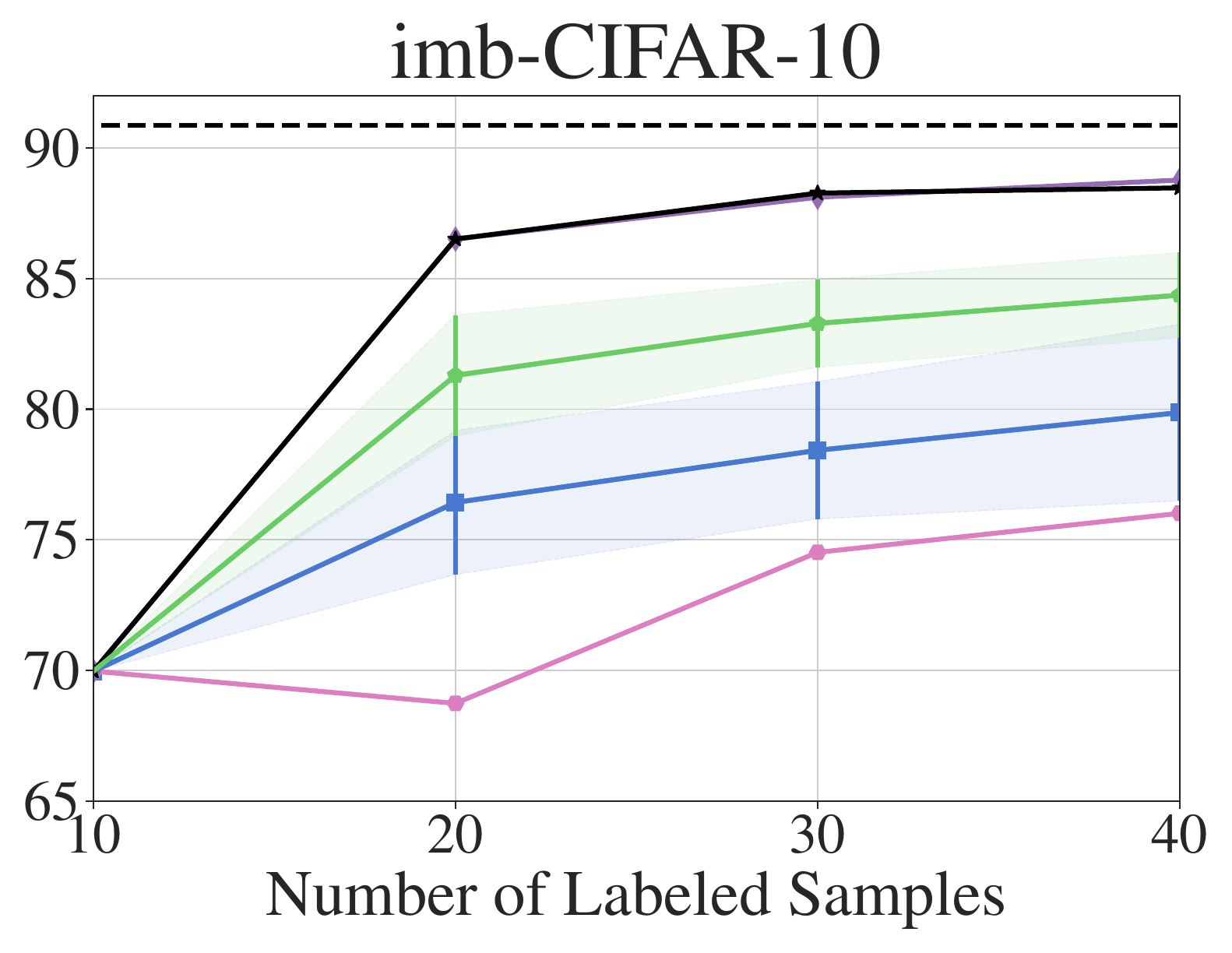}};
    \node[inner sep=0pt] (a3) at (7,0) {\includegraphics[width=3.5cm]{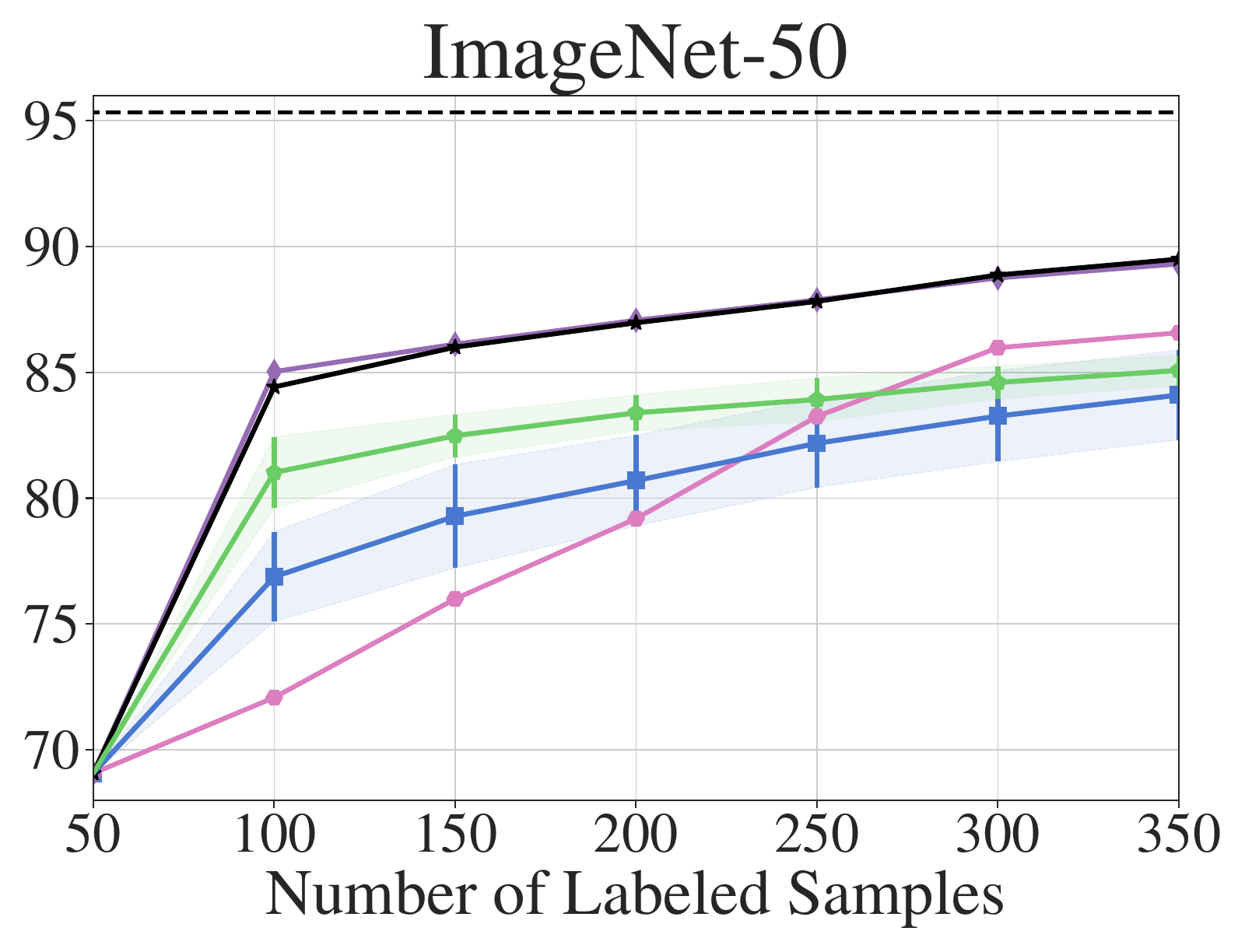}};
    \node[inner sep=0pt] (a4) at (10.5,0) {\includegraphics[width=3.5cm]{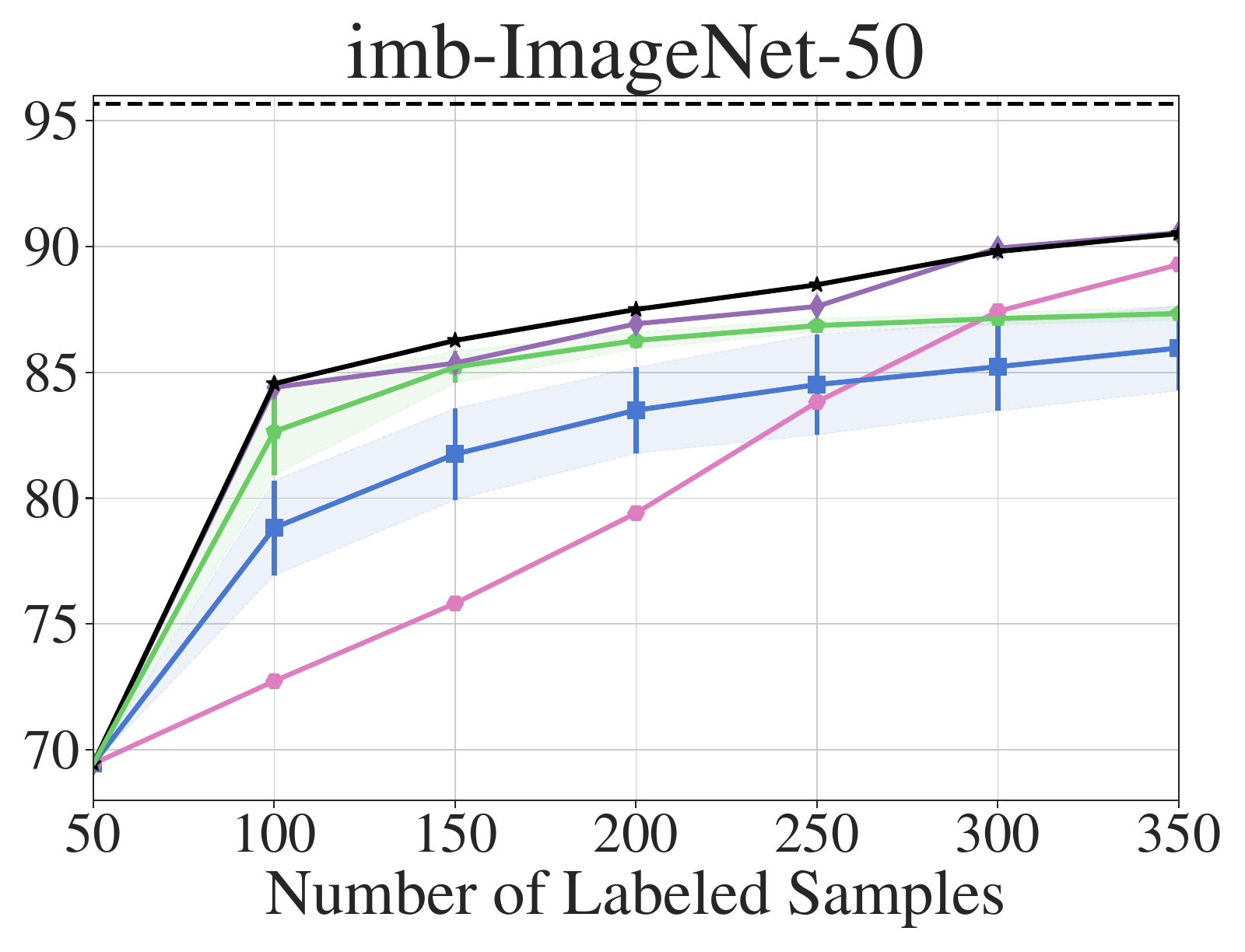}};
    \node[inner sep=0pt] (b1) at (0,-2.8) {\includegraphics[width=3.5cm]{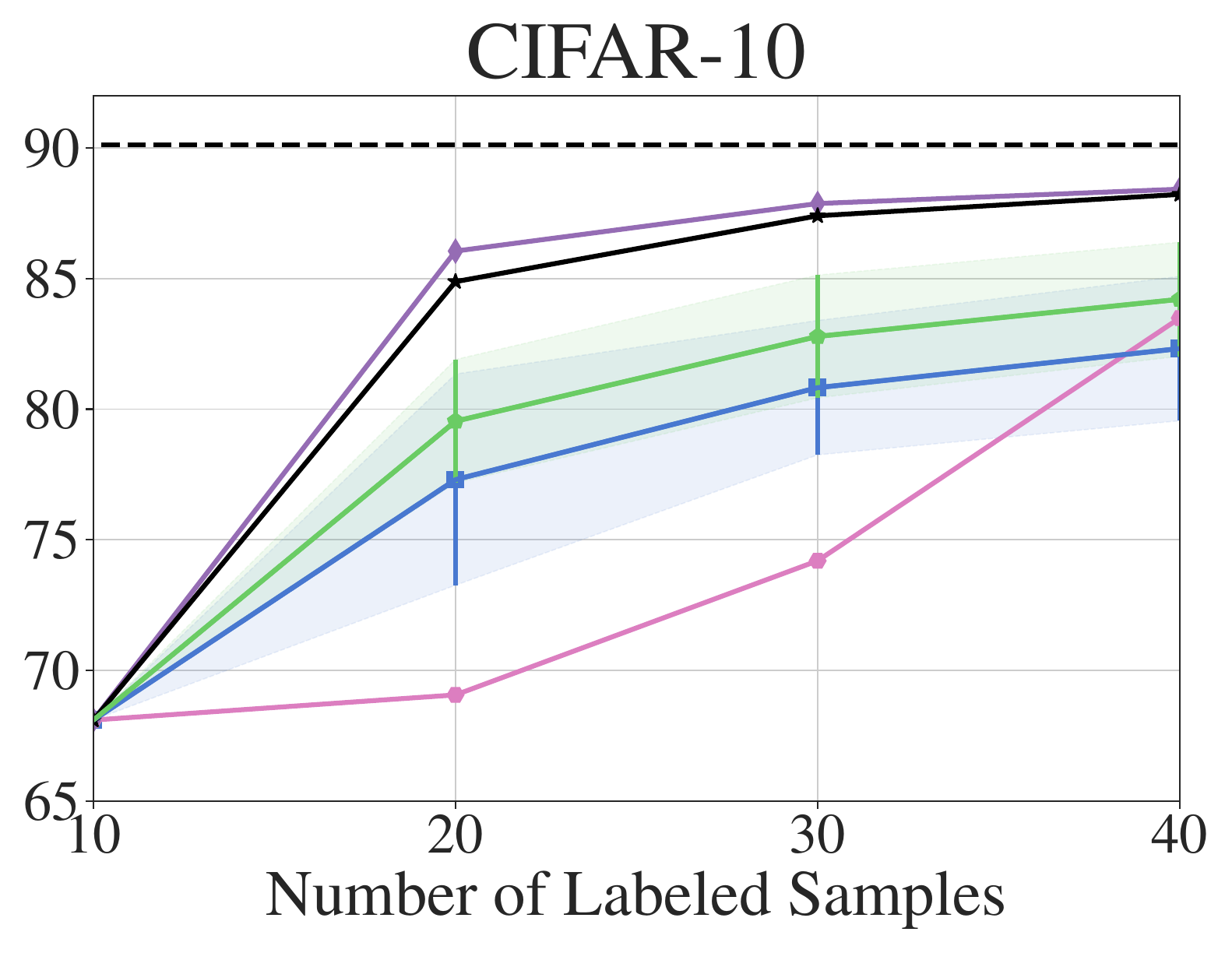}};
    \node[inner sep=0pt] (b2) at (3.5,-2.8) {\includegraphics[width=3.5cm]{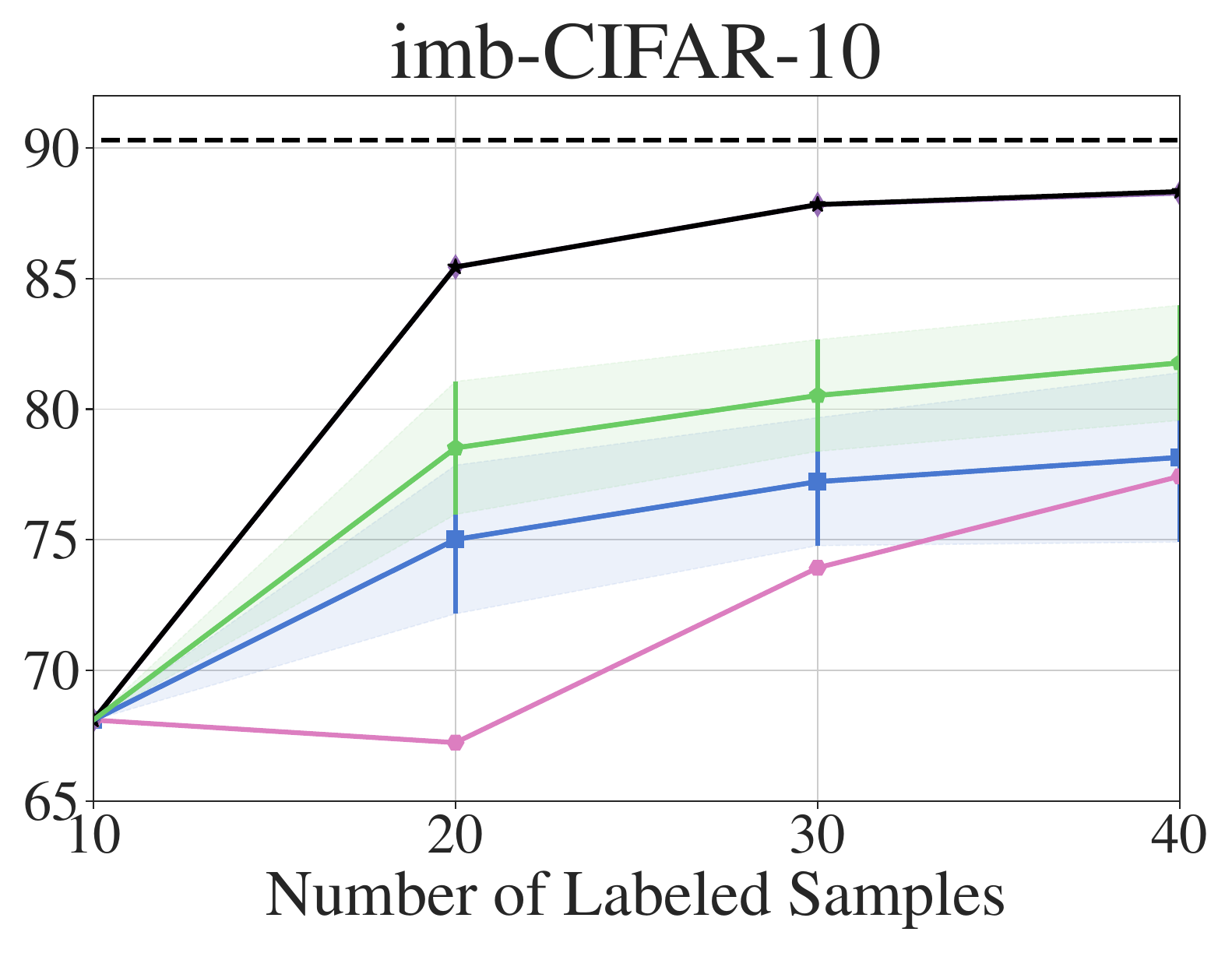}};
    \node[inner sep=0pt] (b3) at (7, -2.8) {\includegraphics[width=3.5cm]{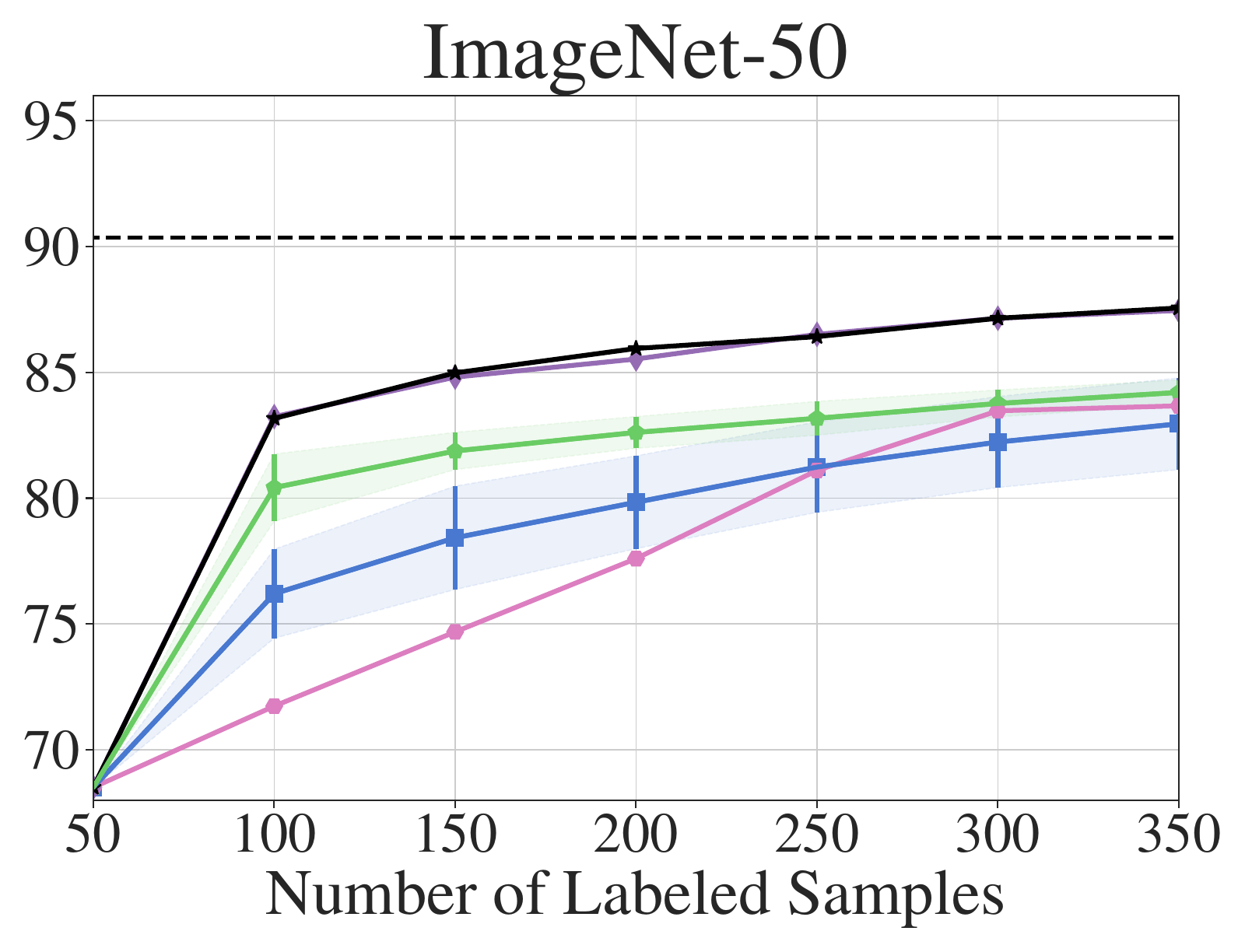}};
    \node[inner sep=0pt] (b4) at (10.5, -2.8) {\includegraphics[width=3.5cm]{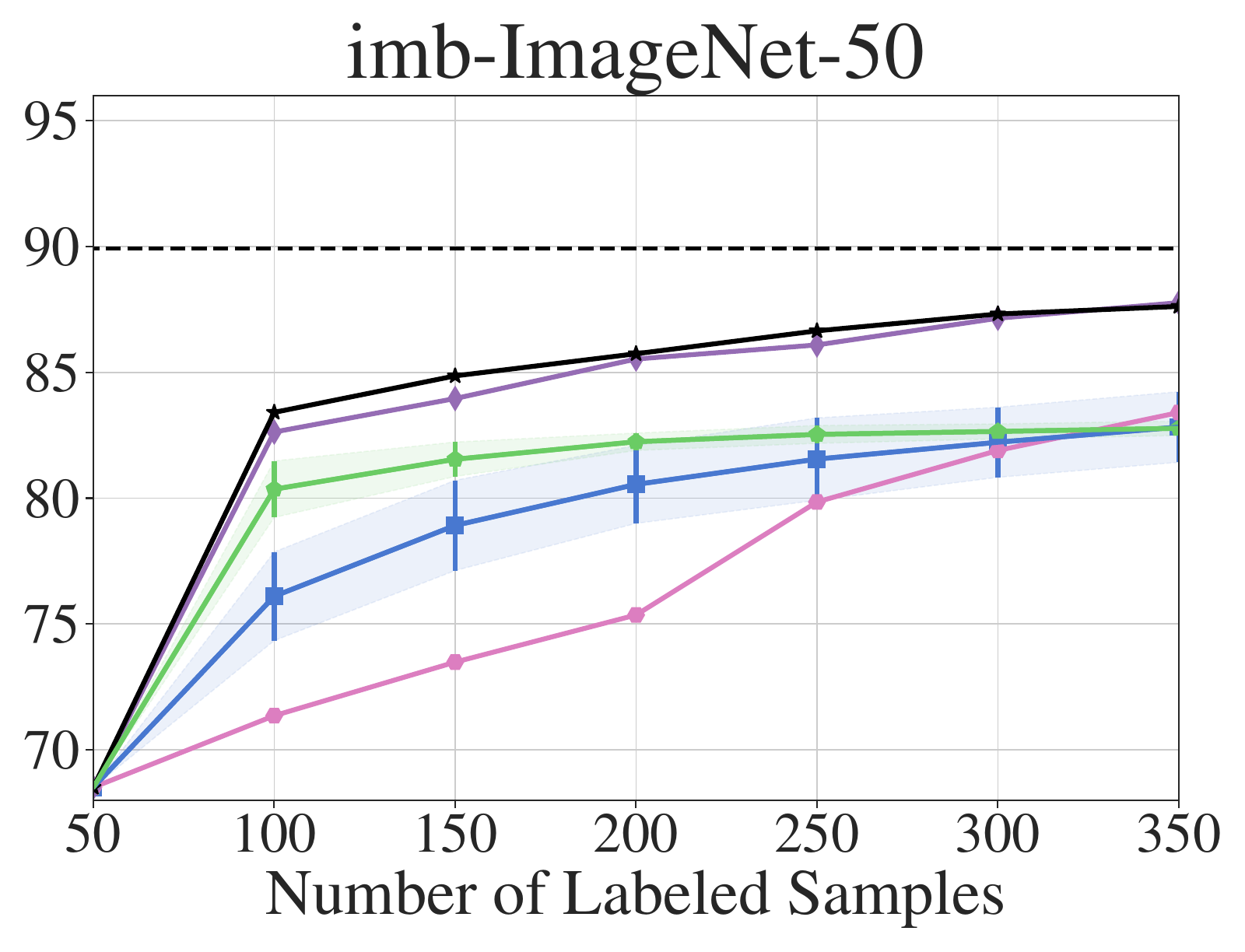}};
\node [anchor=south east,] at (-4.7, 1.0) {\footnotesize (A)};
\node [anchor=south east,] at (-1.2, 1.0) {\footnotesize (B)};
\node [anchor=south east,] at (2.3, 1.0) {\footnotesize  (C)};
\node [anchor=south east,] at (5.8, 1.0) {\footnotesize (D)};
\node [anchor=south east,] at (9.3, 1.0) {\footnotesize (E)};
\node [anchor=south east,] at (-4.7, -1.8) {\footnotesize  (F)};
\node [anchor=south east,] at (-1.2, -1.8) {\footnotesize  (G)};
\node [anchor=south east,] at (2.3, -1.8) {\footnotesize  (H)};
\node [anchor=south east,] at (5.8, -1.8) {\footnotesize  (I)};
\node [anchor=south east,] at (9.3, -1.8) {\footnotesize (J)};
    \end{tikzpicture}
\caption{Classification accuracy for active learning experiments conducted on MNIST, CIFAR-10, imb-CIFAR-10, ImageNet-50, and imb-ImageNet-50 on MNIST, CIFAR-10, imb-CIFAR-10, ImageNet-50 and imb-ImageNet-50. The upper row ((A)-(E)) are plots of \poolacc on the unlabeled pool $\Xu$, the lower row ((F)-(J)) are plots of \evalacc on the evaluation data.}
\label{fig:acc-1}
\end{figure*}


In our accuracy experiments, we attempt to answer the following questions regarding  \nfiral. How does the performance of \nfiral in active learning tests compare to \exactfiral? How does \nfiral compare to other active learning methods? Considering we utilize the Hutchinson trace estimator and CG for gradient estimation in \Relax, what impact do variations in the number of Rademacher random vectors and CG termination criteria have on the convergence of \Relax? \par\medskip

\noindent\textbf{Datasets.} We demonstrate the effectiveness of \nfiral using the following real-world datasets: MNIST~\cite{deng-2012mnist}, CIFAR-10~\cite{cifar10}, Caltech-101~\cite{caltech101} and ImageNet~\cite{imagenet}. First we use unsupervised learning to extract features and then apply active learning to the feature space, that is, we do \textbf{not} use any label information in our pre-processing. For MNIST, we calculate the normalized Laplacian of the training data and use the spectral subspace of the 20 smallest eigenvalues. For CIFAR-10, we use a contrastive learning SimCLR model~\cite{simclr} to extract feature; then we compute the normalized  Laplacian and select the subspace of the 20 smallest eigenvalues. For Caltech-101 and ImageNet-1k, we use state-of-the-art self-supervised learning model DINOv2~\cite{dinov2} to extract features. We additionally select 50 classes randomly from ImageNet-1k and construct dataset ImageNet-50.

We construct 7 datasets for the active learning tests. A summary of the datasets is outlined in \cref{table:acc-data}. For the initial labeled set $\Xo$, we randomly pick two samples per class for ImageNet-1k and one per class for all other datasets. To form the unlabeled pool $\Xu$ in MNIST, CIFAR-10, ImageNet-50, and ImageNet-1k, we evenly select points from each class randomly. To simulate a non-i.i.d. scenario, we assemble $\Xu$ in an imbalanced manner for imb-CIFAR-10, imb-ImageNet-50, and Caltech-101. In imb-CIFAR-10 and Caltech-101, the maximum ratio of points between two classes is 10. In imb-ImageNet-50, the maximum ratio of points between two classes is eight. We use the points from the whole training dataset for evaluation.
\par\medskip

\noindent\textbf{Experimental setup.} 
We compare our proposed \nfiral with four methods: (1) Random selection, (2) K-means where $k=b$ ($b$ is the budget of the active learning selection per round), (3) Entropy: select top-$b$ points that minimize $\sum_c p(y=c|x)\log p(y=c|x)$, (4) \exactfiral: the original implementation of \cref{algo:exact-firal}. For tests involving larger dimension and number of classes, such as Caltech-101 and ImageNet-1k, we do not conduct tests on \exactfiral due to its demanding storage and computational requirements. 

For each of our active learning tests, we use a fixed budget number for selecting points across 3 to 6 rounds. The details are outlined in \cref{table:acc-data}. We report the average and standard deviation for Random and K-means based on 10 trials.

Regarding the hyperparameters in \Relax, we fix the number of Rademacher vectors at 10, and terminate the CG iteration when the relative residual falls below 0.1. Additionally, we conclude the mirror descent iteration when the relative change of the objective is less than \num{1.e-4}. In all of our tests in \cref{table:acc-data}, this criterion is met within fewer than 100 mirror descent iterations. 

The \Round step requires only one hyperparameter, $\eta$. We determine the value of $\eta$ following the same approach as \exactfiral \cite{firal-neurips}: we execute the \Round step with different $\eta$ values, and then select the one that maximizes  $\min_{k\in[c]}\lambda_{\min}(\mathbf{H})_k$, where $\mathbf{H}$ represents the summation of Hessian of the selected $b$ points (\cref{algo:approx-firal}).

We utilize the logistic regression implementation of \texttt{scikit-learn}~\cite{scikit-learn} as our classifier, and we keep the parameters fixed during active learning.

\par\medskip

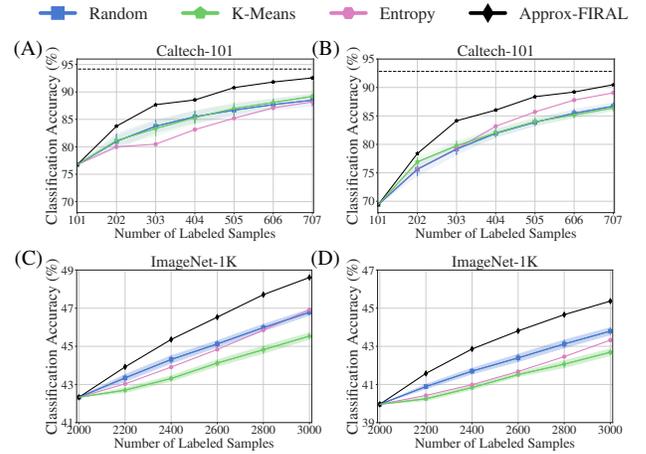
\begin{figure}[!t]
\centering
\begin{tikzpicture}
\node[inner sep=0pt] (c) at (2.,1.7) {\includegraphics[width=8cm]{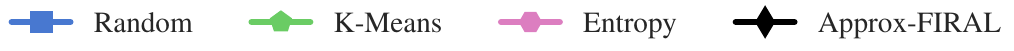}};
    \node[inner sep=0pt] (a1) at (0,0) {\includegraphics[width=3.8cm]{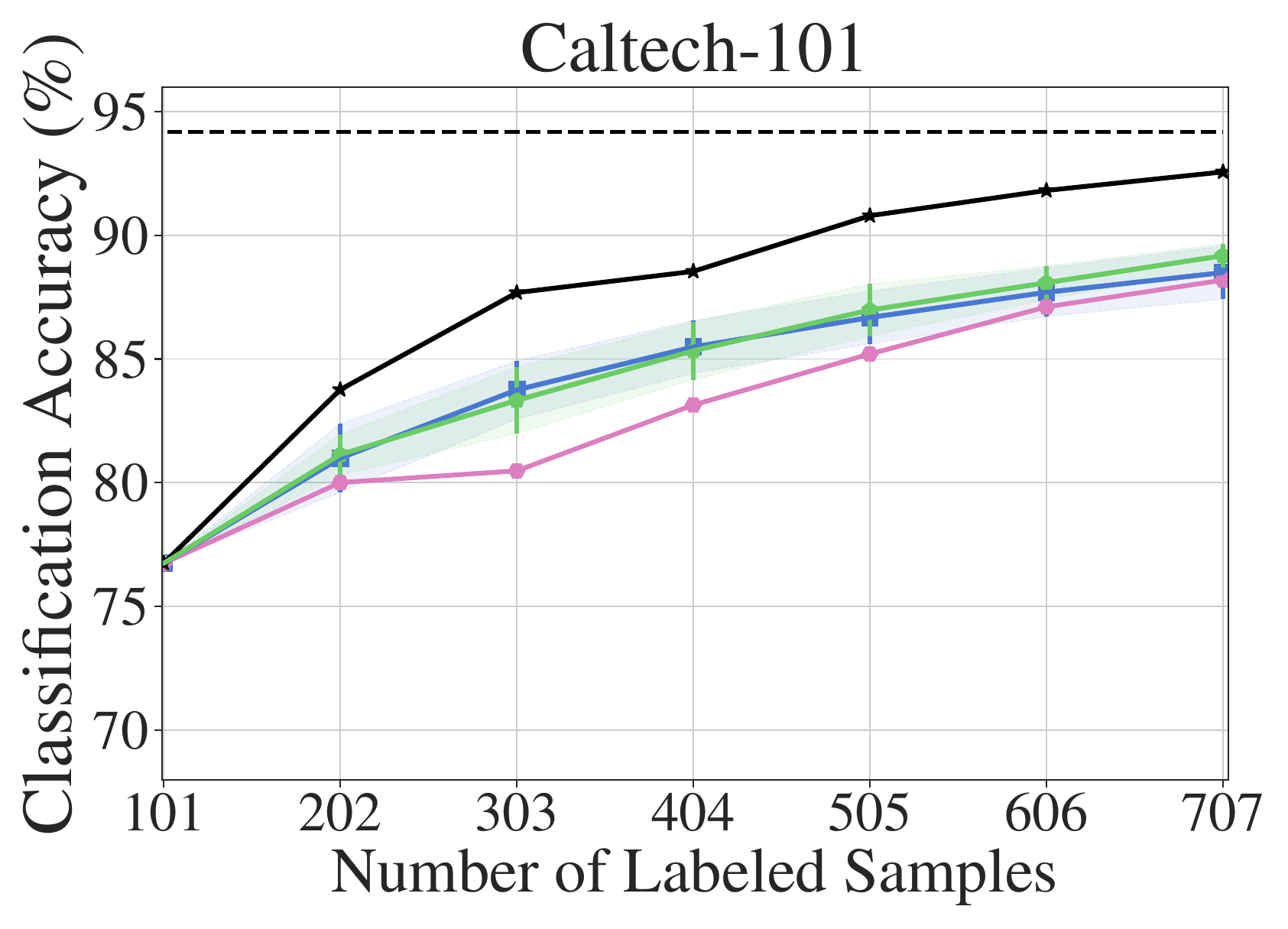}};
    \node[inner sep=0pt] (a2) at (4.,0) {\includegraphics[width=3.8cm]{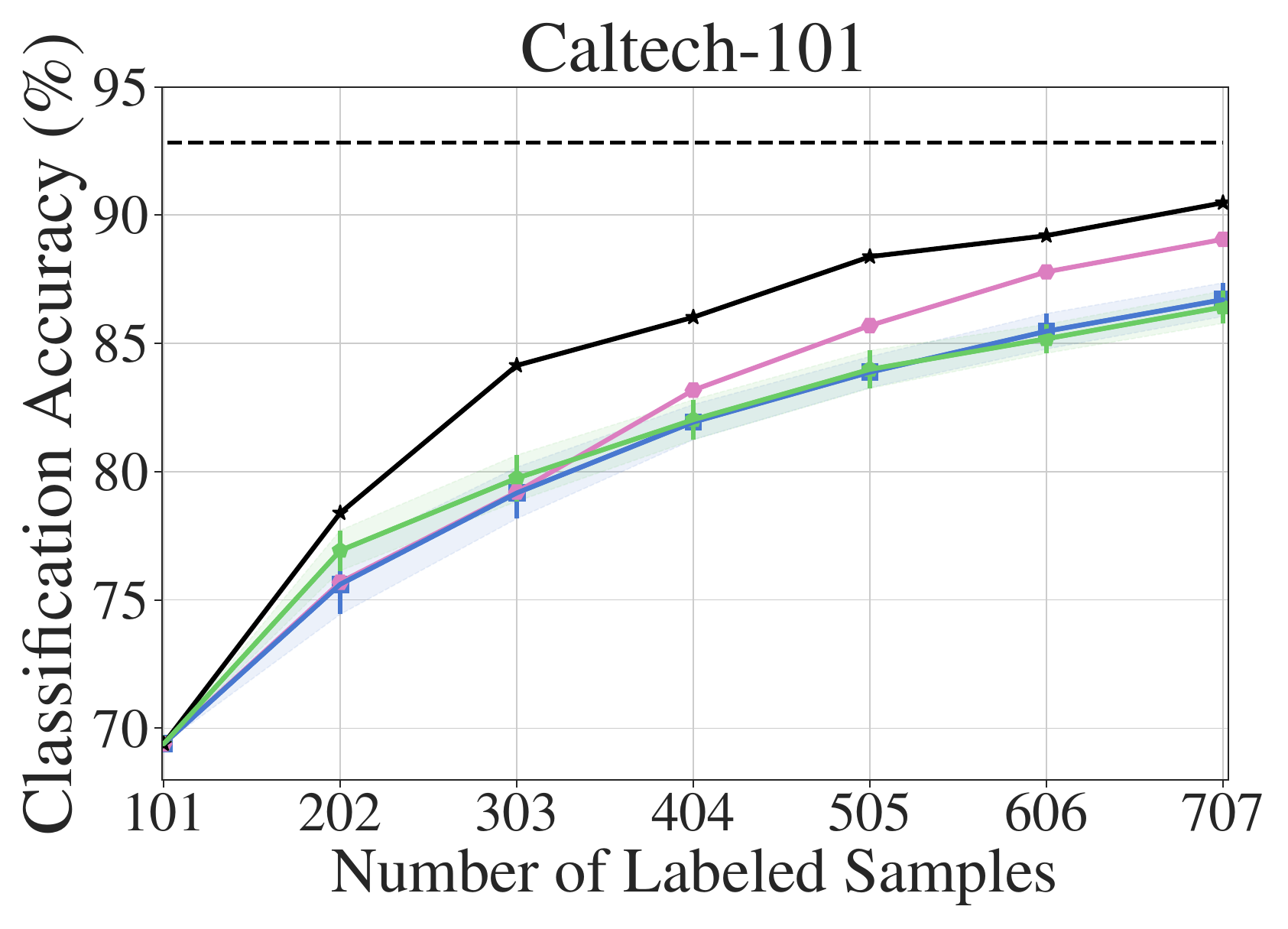}};
    \node[inner sep=0pt] (b1) at (0,-2.8) {\includegraphics[width=3.8cm]{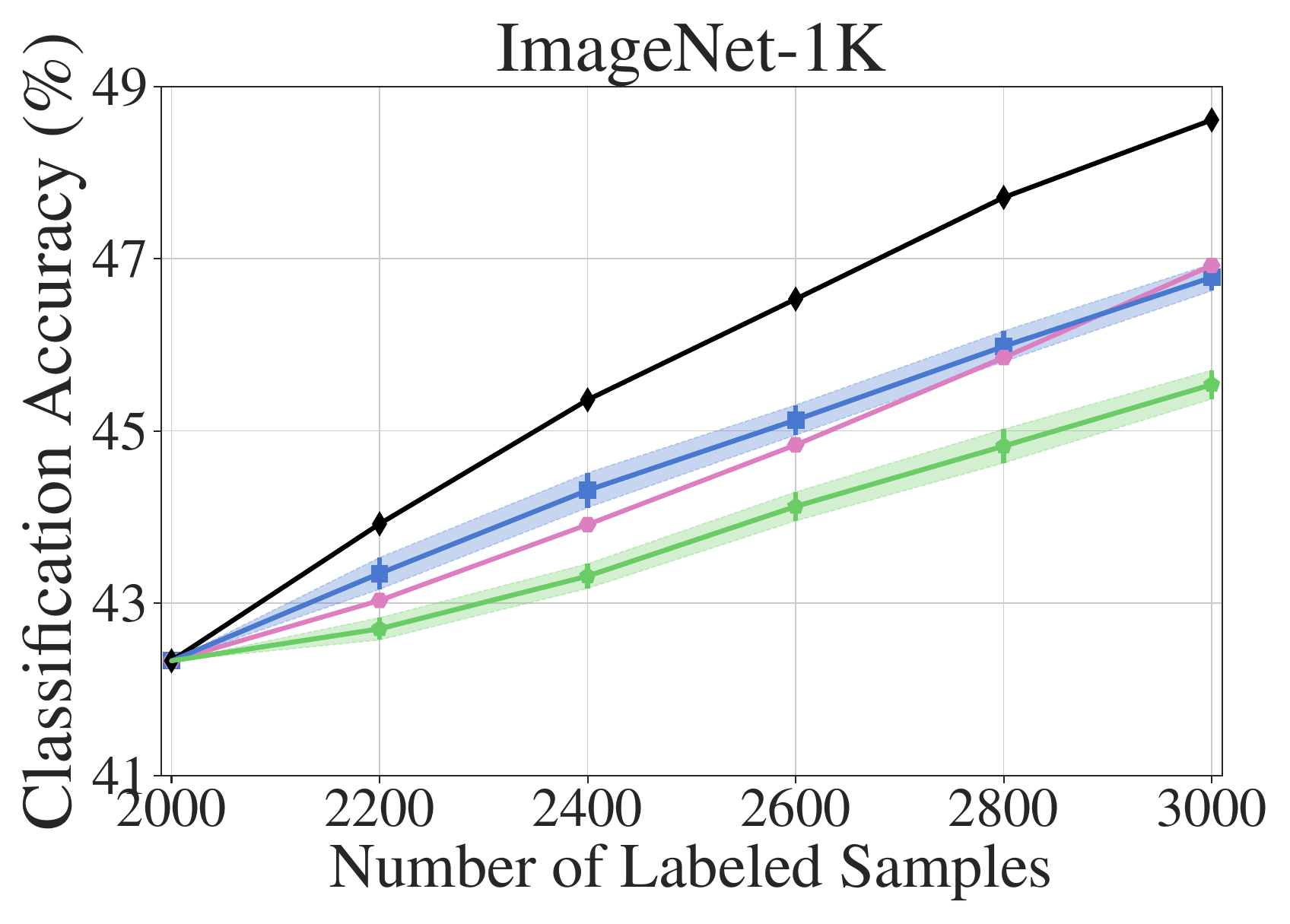}};
    \node[inner sep=0pt] (b2) at (4,-2.8) {\includegraphics[width=3.8cm]{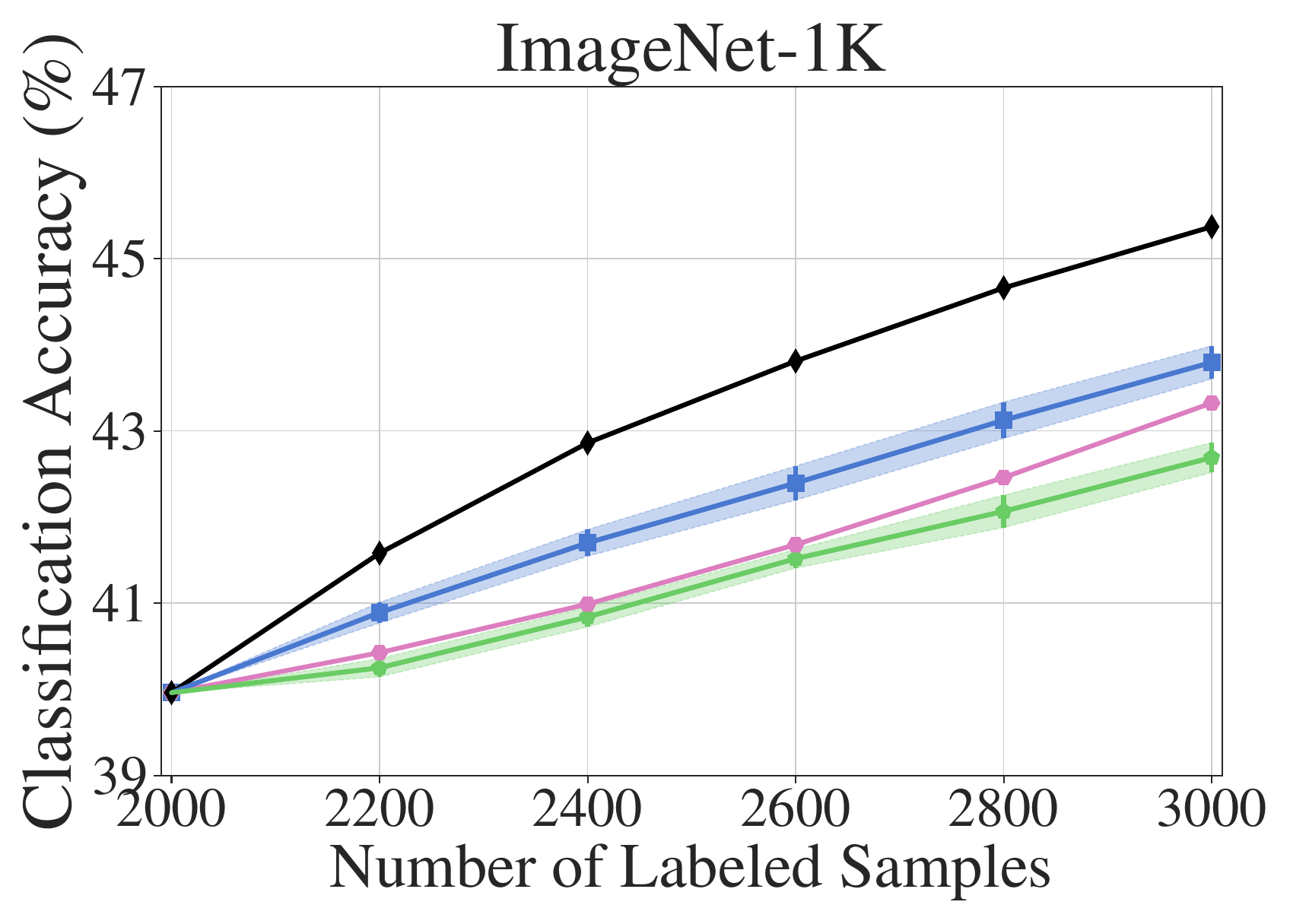}};
\node [anchor=south east,] at (-1.75, 1.0) {\footnotesize (A)};
\node [anchor=south east,] at (2.2, 1.0) {\footnotesize (B)};
\node [anchor=south east,] at (-1.75, -1.8) {\footnotesize (C)};
\node [anchor=south east,] at (2.2,-1.8) {\footnotesize (D)};
\end{tikzpicture}
\caption{Classification accuracy for active learning experiments on Caltech-101 and ImageNet-1k. Both (A) and (B) represent the accuracy on evaluation data for Caltech-101. In (A), the accuracy is averaged with each point having the same weight, while in (B), the accuracy is averaged with each class having the same weight. (C) presents the \poolacc for ImageNet-1k, and (D) presents the \evalacc for ImageNet-1k. }
\label{fig:acc-2}
\end{figure}

We present the classification accuracy results for both \poolacc and \evalacc on MNIST, CIFAR-10, imb-CIFAR-10, ImageNet-50 and imb-ImageNet-50 in \cref{fig:acc-1}. Here, \poolacc refers the accuracy of classifier on the unlabeled pool points $\Xu$, while \evalacc represents the accuracy on the evaluation data  (the respective quantities are detailed in \cref{table:acc-data}). In \cref{fig:acc-2}, we plot the accuracy results obtained from active learning tests conducted on Caltech-101 and ImageNet-1k. \par\medskip

\noindent\textbf{\nfiral vs. \exactfiral.} 
From the results depicted in \cref{fig:acc-1}, we can observe a very close resemblance in the performance of \nfiral and \exactfiral. The discrepancies between these two methods are only visible in a few instances. For example, in the initial round of the CIFAR-10 test (where the number of labeled points is 20 in \cref{fig:acc-1}(B) and (G)), \exactfiral exhibits slightly better performance than \nfiral. However, \nfiral surpasses \exactfiral slightly in the imb-ImageNet-50 test (\cref{fig:acc-1}(E) and (J)), as well as in the final round of the MNIST test (\cref{fig:acc-1}(A) and (F)).

In \cref{table:acc-time}, we illustrate the time comparison between \exactfiral and \nfiral for the initial round of ImageNet-50 and Caltech-101 on a single A100 GPU. For ImageNet-50, \nfiral demonstrates approximately 29 times faster performance than \exactfiral. In the case of Caltech-101, \nfiral is about 177 times faster  compared to \exactfiral.

\par\medskip

\noindent\textbf{\nfiral vs. other methods.}
It is evident that \nfiral outperforms other methods in the active learning test results presented in \cref{fig:acc-1,fig:acc-2}.  Notably, methods such as Random, K-means, and Entropy exhibit an obvious decrease in \evalacc from the balanced CIFAR-10 test (\cref{fig:acc-1}(G)) to the imbalanced CIFAR-10 test (\cref{fig:acc-1}(H)). However, FIRAL maintains a consistent performance level across both CIFAR-10 and imb-CIFAR-10 tests. Further observations include: K-means outperforms Random in all the active learning tests presented in \cref{fig:acc-1}, shows comparable accuracy results to Random in Caltech-101 (\cref{fig:acc-2}(A) and (B)), and exhibits inferior performance compared to Random in ImageNet-1k (\cref{fig:acc-2}(C) and (D)). Additionally, in scenarios where the number of labeled samples is limited (such as in tests on MNIST, CIFAR-10, or the initial rounds of ImageNet-50 in \cref{fig:acc-1}),  Random and K-means display considerable variance, and uncertainty-based method such as Entropy performs the poorest.

\par\medskip

\begin{figure}[!t]
\centering
\begin{tikzpicture}
    \node[inner sep=0pt] (a1) at (0,0) {\includegraphics[width=3.8cm]{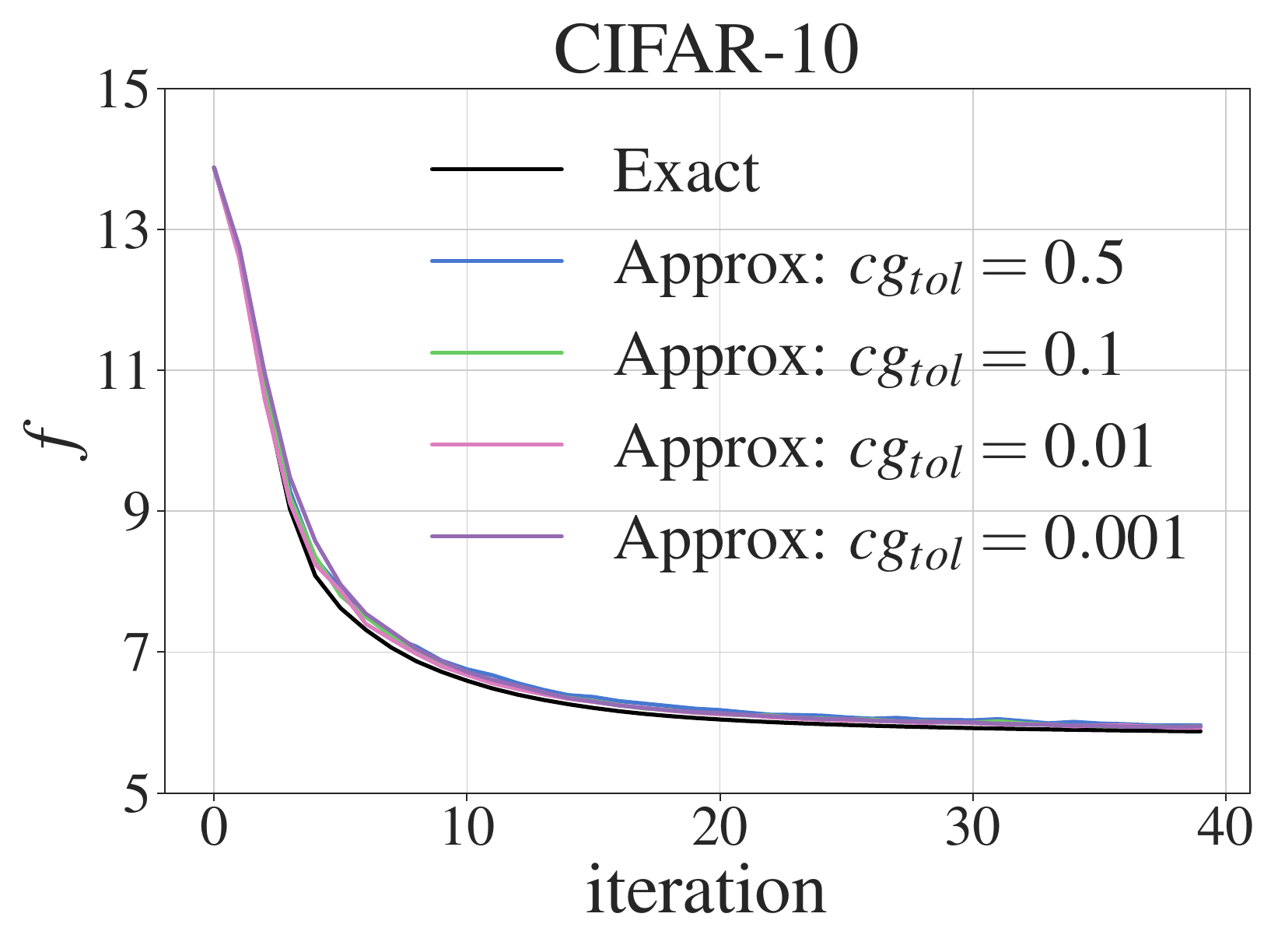}};
    \node[inner sep=0pt] (a2) at (4.,0) {\includegraphics[width=3.8cm]{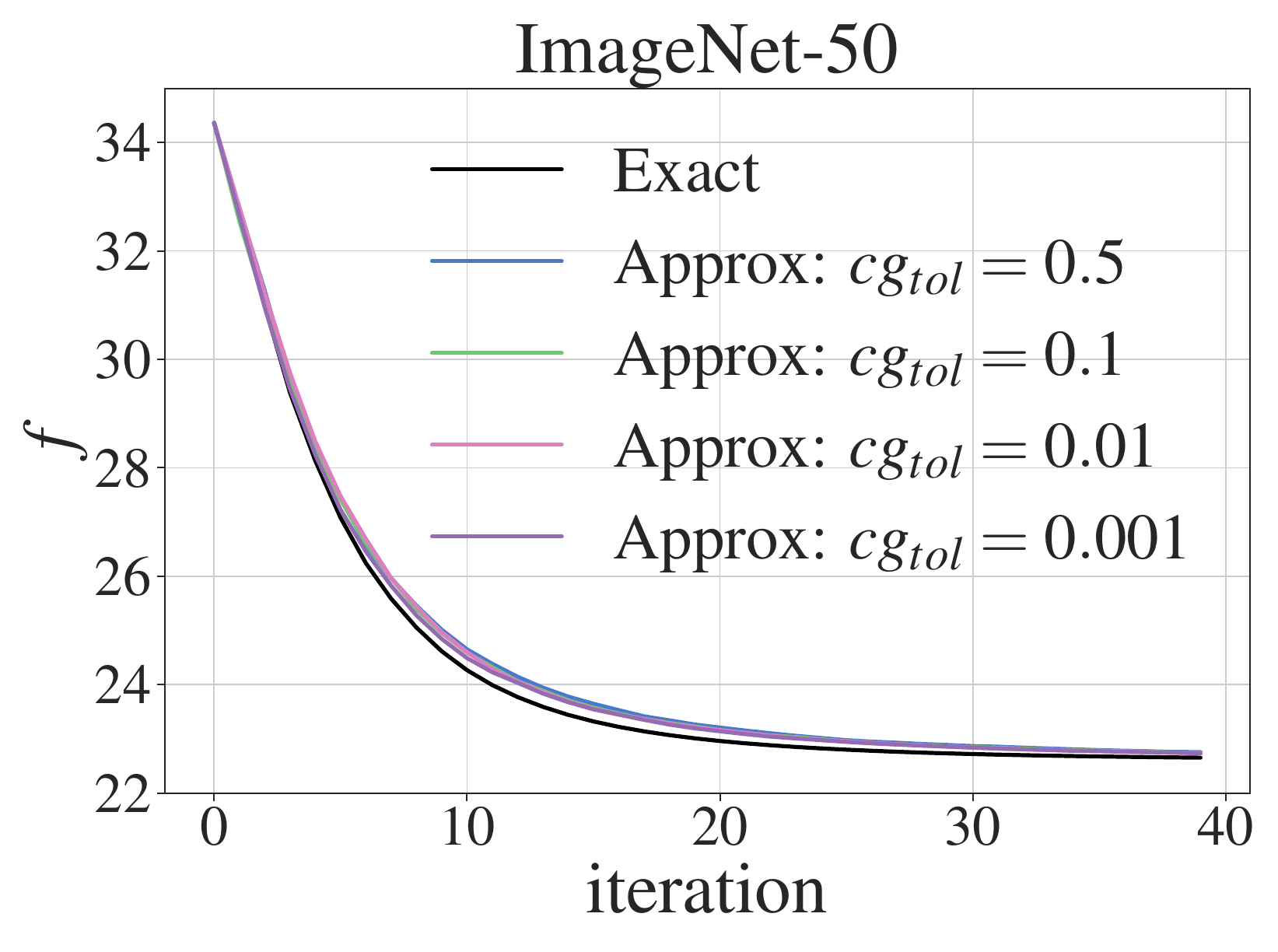}};
    \node[inner sep=0pt] (b1) at (0,-2.8) {\includegraphics[width=3.8cm]{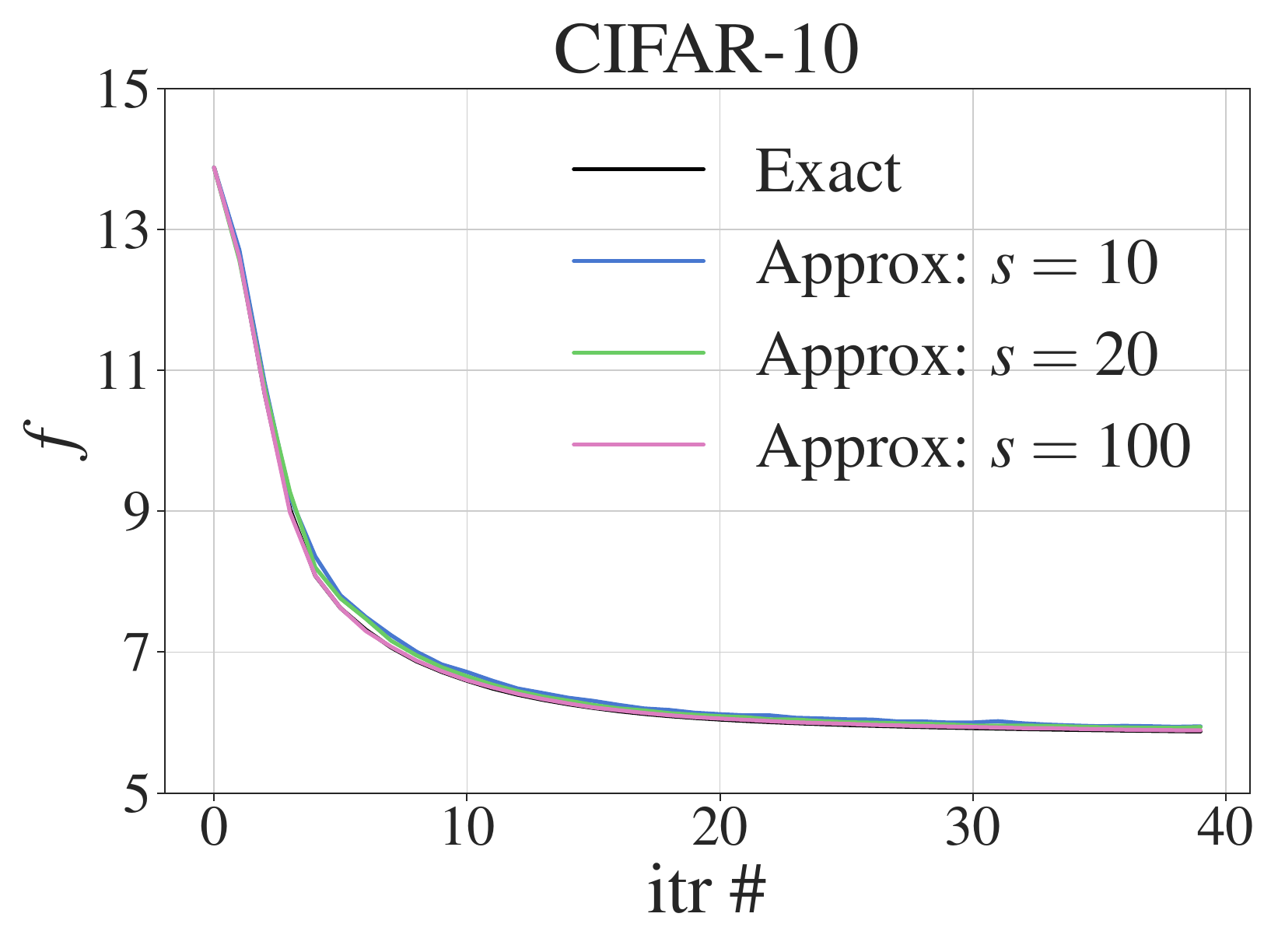}};
    \node[inner sep=0pt] (b2) at (4,-2.8) {\includegraphics[width=3.8cm]{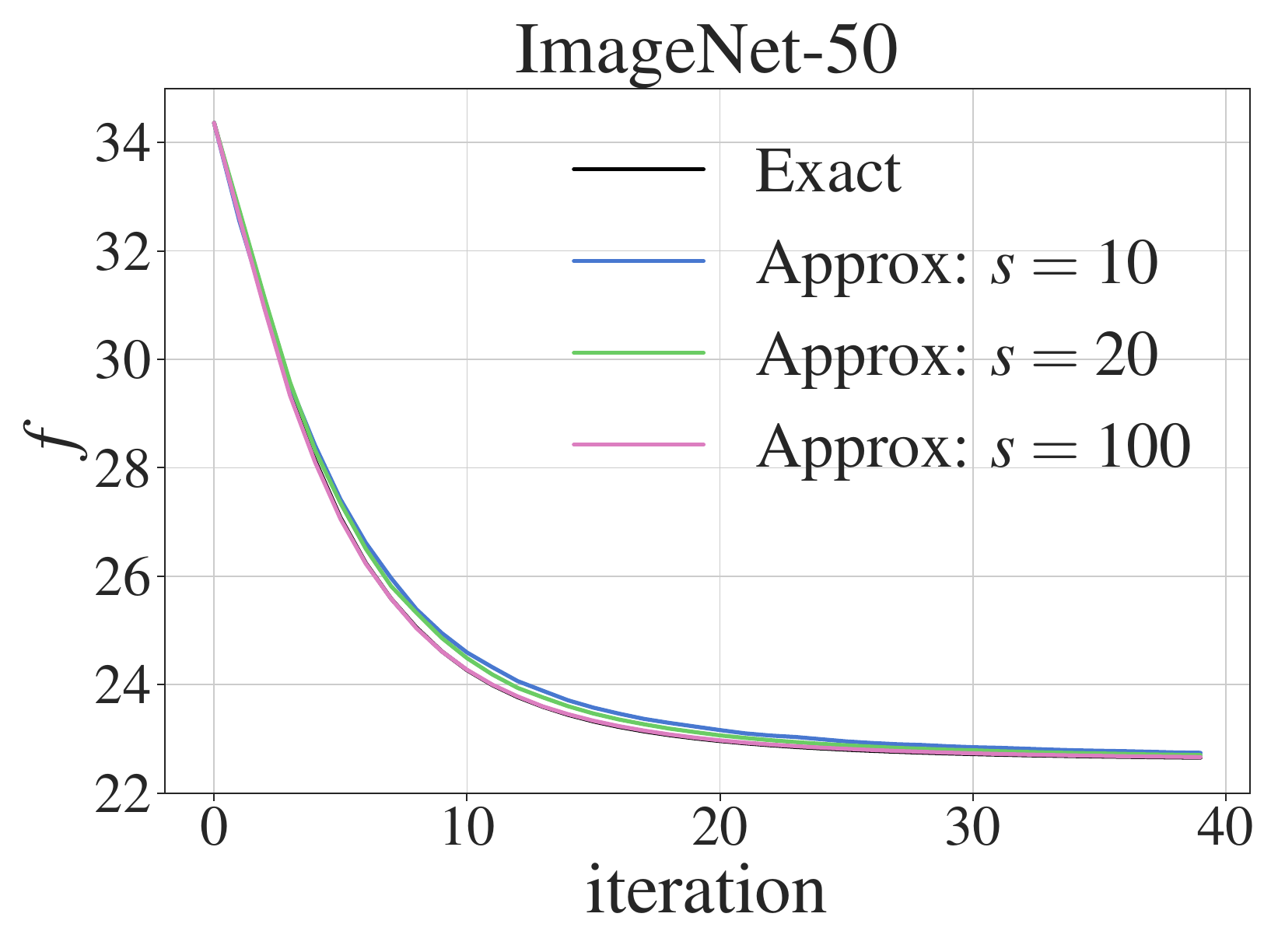}};
\end{tikzpicture}
\caption{Effect of the number of Rademacher random vectors (top) and CG termination criteria (bottom) on \Relax step (i.e., \cref{algo:new_relax}). ``Exact" refers to the precise \Relax solver utilized in \exactfiral, while ``Approx" denotes the fast \Relax solver employed in \nfiral. Here, $s$ denotes the number of Rademacher random vectors, and $cg_{tol}$ signifies the relative residual termination tolerance used in the CG solves.}
\label{fig:relax-sensitivity}
\end{figure}

\noindent\textbf{Parameters in \Relax step.} 
To explore the influence of the number of Rademacher random vectors ($s$) and the termination tolerance of CG ($cg\_tol$) on \Relax, we analyze the initial round of the active learning test on CIFAR-10 and ImageNet-50. We plot the objective function value \cref{eq:obj-relax} of \Relax against the iteration number of mirror descent in \cref{fig:relax-sensitivity}, varying the values of $s$ or $cg\_tol$. Notably, we observe that \Relax does not demonstrate sensitivity to either $s$ or $cg\_tol$.


\begin{figure*}
\centering
\begin{tikzpicture}

\draw (0, 0) node[inner sep=0] {\includegraphics[width=1\textwidth]{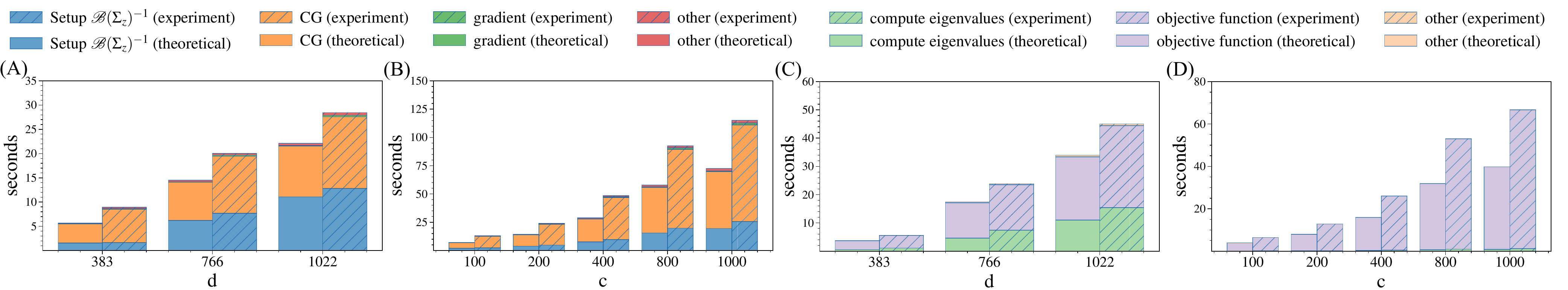}};
\draw (-4, 1.8) node {\footnotesize \Relax solve, ImageNet-1K, single-node};
\draw (4, 1.8) node {\footnotesize \Round solve, ImageNet-1K, single-node};

\end{tikzpicture}
\caption{Wall-clock time dependence of the \Relax and \Round solves to the number of features $d$ and the number of classes $c$ using  ImageNet-1K. In the run for  the $d$ scaling, we fix the number of data points $n=\num{1E5}$ and the number of classes $c=1000$. We set the number of random vectors to $s=10$. For each value of $d$, we run one gradient and fix the number of CG iterations to $n_{CG}=50$; and the left column represents theoretical time and the right column represents experimental time. In the run to test the algorithmics scalability in $c$, we fix $n=\num{1.3E6}$, $d=383$ and vary c as $\left[100, 200, 400, 800, 1000\right]$ . The remaining parameters of the algorithm are fixed. We report the results as follows{(A)} {\Relax} run for $d$ scaling. {(B)} {\Relax} run for $c$ scaling. {(C)} {\Round} solve for $d$ scaling. {(D)} {\Round} solve for $c$ scaling.}
\label{fig: sensitivity}
\end{figure*}

\begin{figure*}
\centering
\includegraphics[width=1\textwidth]{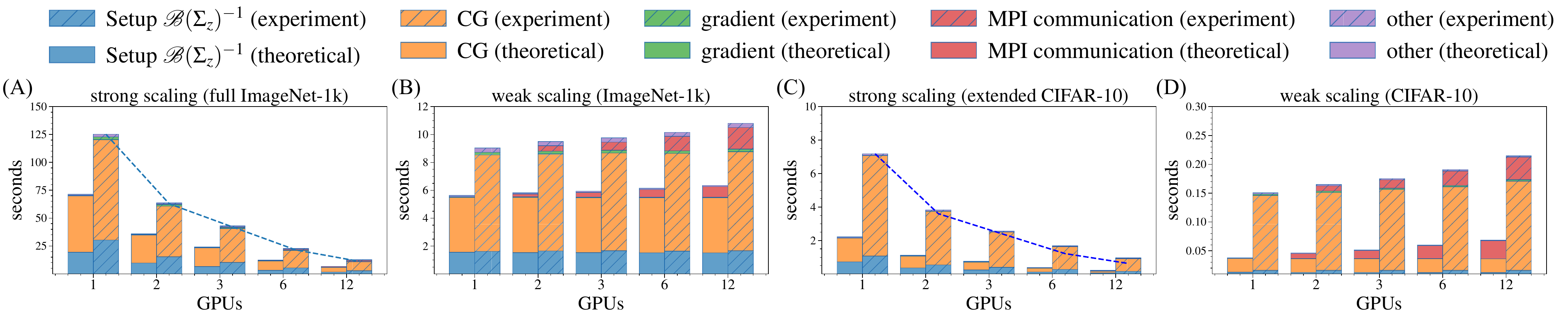}
\caption{Strong and weak scaling of the \Relax step on CIFAR-10 and ImageNet-1K. The dashed lines indicate ideal scaling performance. (A) Strong scaling on the full ImageNet-1K dataset (\num{1.3E6} points). (B) Weak scaling on ImageNet-1K (\num{1E5} points per rank). (C) Strong scaling on the extended CIFAR-10 dataset (\num{3E6} points). (D) Weak scaling on CIFAR-10 (\num{5E4} points per rank).}
\label{fig: relax_solve_scaling}
\end{figure*}

\begin{figure*}
\centering
\includegraphics[width=1\textwidth]{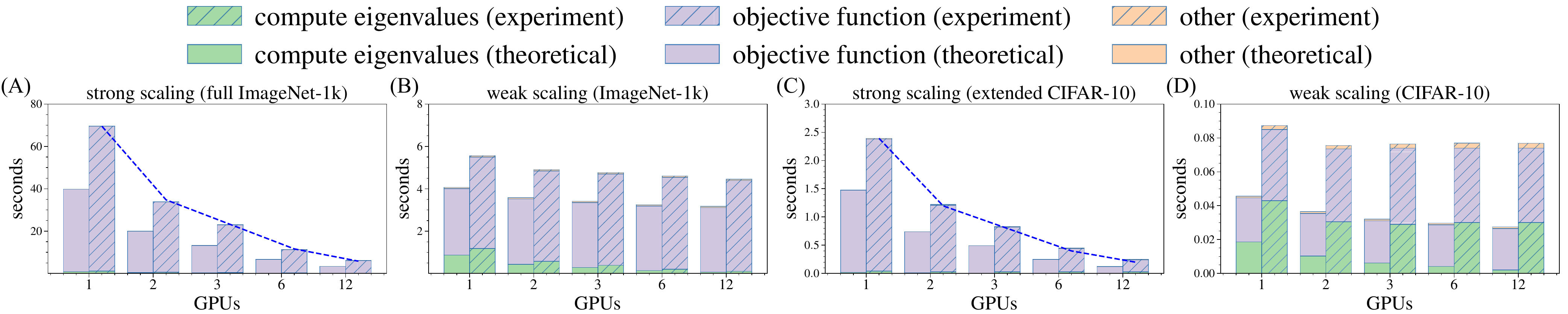}
\caption{Strong and weak scaling of the \Round step on CIFAR-10 and ImageNet-1K. The dashed lines indicate ideal scaling performance. (A) Strong scaling on the full ImageNet-1K dataset (\num{1.3E6} points). (B) Weak scaling on ImageNet-1K (\num{1E5} points per rank). (C) Strong scaling on the extended CIFAR-10 dataset (\num{3E6} points). (D) Weak scaling on CIFAR-10 (\num{5E4} points per rank).}
\label{fig: round_solve_scaling}
\end{figure*}


\subsection{Single-GPU performance}\label{s:result-single}
We now turn our attention to the HPC performance evaluation. We start with discussing  the  performance of our algorithm on a single GPU. We study the performance sensitivity to feature size $d$ and number of class $c$ in ImageNet dataset for both \Relax step and \Round step. We provide estimates for the theoretical peak time of each major computational component, assuming an ideal peak performance of 19.5TFLOPS for Float32 computation on the GPU A100~\cite{nvidia2020}. The computation of \Relax solve is broken down into four major components: setting up preconditioner $\mathcal{B}(\bSigma_z)^{-1}$, performing the conjugate gradient (\textit{CG}), evaluating the \textit{gradient} and \textit{other} related tasks. For \Round solve, we focus on three components: \textit{computing eigenvalues} that is invoked at line 9 of \cref{algo:approx-firal}, evaluating the \textit{objective function}, and \textit{other} related tasks. \par

\noindent\textbf{Sensitivity to feature size $d$.} As we saw, the computational  complexity of the \Relax solve is $\mathcal{O}\left(cd^3 + ncd^2 + n_{CG}ncsd\right)$, where $n_{CG}$ is the number of CG iterations. The major cost lies in the construction of preconditioner $\left\{\mathcal{B}_k(\bSigma_z^{-1})\right\}_{k\in[c]}$ and CG solving $\bSigma_z \b W=\b V$. The construction of $\left\{\mathcal{B}_k(\bSigma_z^{-1})\right\}_{k\in[c]}$ takes $cd^3 + 2cnd^2$ time. The CG solve involves $n_{CG}$ evaluations of $\bSigma_z \b V$. According to \cref{lm:matvec}, the time complexity of CG is dominated by $4n_{CG}ncsd$. Time complexity of \Round solve is $\mathcal{O}(cd^3 + ncd^2)$. The major cost lies in line 9 in \cref{algo:approx-firal}, and evaluation of objective function in \cref{eq:obj-round-sparse}. We use the \texttt{cupy.linalg.eigvalsh} to compute eigenvalues which takes $\mathcal{O}(cd^3)$. We fit the prefactor to 300 doing a few experiments isolated to this function. The evaluation of \cref{eq:obj-round-sparse} has time complexity $3cd^3 + 4ncd^2$.
We utilize features from ImageNet-1K extracted using the pretrained self-supervised ViT models DINOv2~\cite{dinov2} with varying dimensions. Specifically, we explore feature dimensions $d$ of 383, 766, and 1022. The number of classes is fixed at $c=1000$, and we maintain a consistent number of points at 500,000.
We set the number of random vectors $s=10$ in $\bold{V}$. \cref{fig: sensitivity}{(A)(C)} show the sensitivity results for both \Relax and \Round steps. Each $d$ value is represented by two adjacent columns. The left column displays the theoretical peak time for each $d$, while the right column shows the actual test time. Specifically, \cref{fig: sensitivity} {(A)} presents the results for the \Relax step. We conduct the \Relax step for one mirror descent iteration while keeping the number of CG iterations fixed at $n_{CG}=50$. Increasing $d$ from 383 to 766 leads to a 4.72$\times$ increase in the wall time of the preconditioner $\{\mathcal{B}_k(\bSigma_z^{-1})\}_{k\in[c]}$, while the CG time increases by 1.7$\times$. When $d$ increases from 766 to 1022, the wall time of the preconditioner increases by 1.66$\times$, and the CG time increases by 1.26$\times$.

In the \Round step, we conduct one iteration and showcase the results in \cref{fig: sensitivity}{(C)}. Increasing $d$ from 383 to 766 results in a 6.6$\times$ increase in eigenvalue computation time. Additionally, the evaluation time for the objective function increases by 3.65$\times$. Upon further increasing $d$ from 766 to 1022, the time required for eigenvalue decomposition increases by 2.08$\times$, while the evaluation time for the objective function rises by 1.79$\times$. \par\medskip

\noindent\textbf{Sensitivity to class number $c$.} Similarly, we examine the algorithm's sensitivity to the number of classes, $c$. As observed, the complexity of the \Relax step scales linearly with the number of classes, as does the construction of the preconditioner. Similarly, the two primary components of the \Round step, namely, computing eigenvalues (line 9 in \cref{algo:approx-firal}) and evaluating the objective function (line 7 in \cref{algo:approx-firal}), also show linear scale to the number of classes. We conduct tests on the ImageNet dataset with 1.3 million points and a feature dimension of $d=383$. The number of classes $c$ varies from ${100, 200, 400, 800, 1000}$. \cref{fig: sensitivity} {(B)} illustrates the results of the \Relax step. When $c$ increases from 100 to 200, the preconditioner cost increases by 2$\times$, and the CG time increases by 1.79$\times$. Conversely, for the scenario where $c$ increases from 100 to 1000, the preconditioner time increases by 10.6$\times$, and the CG time increases by 8.3$\times$. In the \Round step, we execute one iteration and present the results in \cref{fig: sensitivity} {(D)}. As $c$ increases from 100 to 200, the eigenvalue decomposition time increases by 2.08$\times$, and the time for evaluating the objective function increases by 1.99$\times$. Conversely, when $c$ increases from 100 to 1000, the eigenvalue decomposition time increases by 10$\times$, and the time for evaluating the objective function increases by 10.37$\times$. Overall, the solver exhibits the expected scaling behavior.

\subsection{Parallel scalability}\label{s:result-parallel}
We perform strong and weak scaling tests on our parallel implementation of \nfiral using two datasets. \nbone ImageNet-1K: the dimension of points is $d=383$, and the number of classes is $c=1000$. For the strong scaling test, we use the entire ImageNet-1K dataset with an unlabeled pool $\Xu$ containing $n=1.3$ million points. In the weak scaling test, we allocate 0.1 million points to each GPU. \nbtwo CIFAR-10: the dataset has points with dimension of $d=512$ and number of classes  $c=10$ . In the strong scaling test, we expand CIFAR-10 by introducing random noise from $\sim$50K  to 3 million points. For the weak scaling test, we allocate 50,000 points to each GPU.

We present strong and weak scaling results for the \Relax steps in \cref{fig: relax_solve_scaling} and the \Round step in \cref{fig: round_solve_scaling}, employing up to 12 GPUs for both tests. In the \Relax step, we present the time for one mirror descent iteration. For the \Round step, we report the time for selecting one point. For estimating the theoretical collective communication time costs for MPI operations, we assume a latency of $t_s=\num{1E-4}$s, a bandwidth of $1/t_w=\num{2E10}$ byte/s, and a computation cost per byte of $t_c=\num{1E-10}$ s/byte. Additionally, for computation estimation, we maintain the use of 19.5TFLOPS peak performance of GPU A100 as in the previous section. \par\medskip

\noindent\textbf{Scalability of \Relax step.} The main computational cost in the \Relax step stem from the preconditioner setup and the CG solve. Regarding strong scaling results presented in \cref{fig: relax_solve_scaling}(A) for ImageNet-1K and (C) for CIFAR-10, utilizing 12 GPUs leads to a speedup of 10.9$\times$ for the preconditioner setup and 11.3$\times$ for the CG solve in the case of ImageNet-1K. For CIFAR-10, the speedup for the preconditioner is 6.7$\times$, while for CG, it reaches 8 when employing 12 GPUs.  As for the weak scaling, with the number of GPUs raised to 12, the time increases by less than 10\% for ImageNet-1K (\cref{fig: relax_solve_scaling}(B)), and within 20\% for CIFAR-10 (\cref{fig: relax_solve_scaling}(D)). The primary increases in time are attributed to MPI communications. We present the ideal speedup as dashed lines in \cref{fig: relax_solve_scaling}, with negligible variance in performance.
 \par\medskip

\noindent\textbf{Scalability of \Round step.} 
In the \Round step, communication costs are negligible, so we include the time in the "other" category in the plots of \cref{fig: round_solve_scaling}. In the strong scaling tests, employing 12 GPUs results in an 11.4$\times$ speedup for ImageNet-1K, as shown in \cref{fig: round_solve_scaling}(A), and achieves an 11.1$\times$ speedup for CIFAR-10, as seen in \cref{fig: round_solve_scaling}(C). Regarding weak scaling, the time slightly decreases when we increase the number of GPUs. This occurs because we evenly distribute the eigenvalue calculations across all GPUs. This effect is more pronounced in the case of ImageNet-1K, as shown in \cref{fig: round_solve_scaling}(B), compared to CIFAR-10 (\cref{fig: round_solve_scaling}(D)), since ImageNet-1K has 1000 classes while CIFAR-10 has only 10 classes. Similarly, we present the ideal speedup as dashed lines in \cref{fig: round_solve_scaling}, with negligible variance in performance.

Regarding the discrepancy between theoretical and experimental performance shown in \cref{fig: relax_solve_scaling,fig: round_solve_scaling}, one cause is the performance of \texttt{cupy.einsum}, which is impacted by memory management and suboptimal kernel performance for certain input sizes. Additionally, the theoretical analysis includes certain constants related to specific kernels that have not been calibrated, such as the prefactors in the eigenvalue solvers, contributing to the gap.
\section{Conclusions} \label{s:conclusion}
We presented \nfiral, a new algorithm that is orders of magnitude faster than FIRAL. This improvement is achieved by replacing FIRAL's exact solutions with inexact iterative methods or block diagonal approximations, using randomized approximations for matrix traces, and approximating eigenvalue solves with block diagonal methods. Empirical results show that these approximations have minimal impact on sample selection effectiveness, as demonstrated by test accuracy across seven diverse datasets, including those with class imbalances. Furthermore, our muli-GPU implementation allows efficient scaling to large datasets such as ImageNet.  Our open-source Python implementation allows interoperability with existing machine learning workflows.

Our approach has {\bf several limitations}. First, we still use direct solvers in some parts of the code. Specifically, eigenvalue solves in the \Round step and block factorization for our Hessian preconditioner are performed exactly. These methods are not scalable for certain parameters and could be replaced with sparsely preconditioned iterative solvers to enhance both performance and scalability of \nfiral. We aim to incorporate these improvements in future versions of the algorithm. Second, we have not extended the theoretical results of FIRAL to the approximate version. While most of the matrices involved are symmetric positive definite and our approximations are stable perturbations, deriving precise error bounds requires detailed estimates of the approximation error.Third, despite its efficiency, \nfiral is still more resource-intensive compared to other methods. It performs best when the number of classes is relatively small, the feature embeddings are excellent, and only a few examples are needed for classification. As the number of classes grows, simpler methods may be more appropriate. Fourth, our testing has been limited to NVIDIA GPUs. Although the code is theoretically portable—CuPy supports AMD GPUs and it can be adapted for CPUs using NumPy—these alternative implementations have not yet been carried out.

A final, more fundamental limitation of the generic FIRAL approach is its inability to accommodate changes in the feature embedding as new examples are introduced. Typically, empirical methods address this by retraining or fine-tuning the embedding whenever new labels are obtained. In such cases, since the embedding evolves with new data, the data points themselves change, rendering the FIRAL theory inapplicable. Active learning with theoretical guarantees  for such setups remains an open problem that probably requires an entirely different approach.

\section*{Acknowledgements}

This material is based upon work supported by NSF award OAC 2204226; by the U.S. Department of Energy, Office of Science, Office of Advanced Scientific Computing Research, Applied Mathematics program, Mathematical Multifaceted Integrated Capability Centers (MMICCS) program, under award number DE-SC0023171; by the U.S. Department of Energy, National Nuclear Security Administration Award Number DE-NA0003969; and by the U.S. National Institute on Aging under award number  R21AG074276-01. Any opinions, findings, and conclusions or recommendations expressed herein are those of the authors and do not necessarily reflect the views of the DOE, NIH, and NSF. Computing time on the Texas Advanced Computing Centers Stampede system was provided by an allocation from TACC and the NSF.

\bibliographystyle{IEEEtran}
\bibliography{reference,neurips}

\begin{thebibliography}{10}
\providecommand{\url}[1]{#1}
\csname url@samestyle\endcsname
\providecommand{\newblock}{\relax}
\providecommand{\bibinfo}[2]{#2}
\providecommand{\BIBentrySTDinterwordspacing}{\spaceskip=0pt\relax}
\providecommand{\BIBentryALTinterwordstretchfactor}{4}
\providecommand{\BIBentryALTinterwordspacing}{\spaceskip=\fontdimen2\font plus
\BIBentryALTinterwordstretchfactor\fontdimen3\font minus
  \fontdimen4\font\relax}
\providecommand{\BIBforeignlanguage}[2]{{%
\expandafter\ifx\csname l@#1\endcsname\relax
\typeout{** WARNING: IEEEtran.bst: No hyphenation pattern has been}%
\typeout{** loaded for the language `#1'. Using the pattern for}%
\typeout{** the default language instead.}%
\else
\language=\csname l@#1\endcsname
\fi
#2}}
\providecommand{\BIBdecl}{\relax}
\BIBdecl

\bibitem{bengio2013}
Y.~Bengio, A.~Courville, and P.~Vincent, ``Representation learning: A review
  and new perspectives,'' \emph{IEEE transactions on pattern analysis and
  machine intelligence}, vol.~35, no.~8, pp. 1798--1828, 2013.

\bibitem{simclr}
\BIBentryALTinterwordspacing
T.~Chen, S.~Kornblith, M.~Norouzi, and G.~E. Hinton, ``A simple framework for
  contrastive learning of visual representations,'' 2020. [Online]. Available:
  \url{https://arxiv.org/abs/2002.05709}
\BIBentrySTDinterwordspacing

\bibitem{zhuang2020}
F.~Zhuang, Z.~Qi, K.~Duan, D.~Xi, Y.~Zhu, H.~Zhu, H.~Xiong, and Q.~He, ``A
  comprehensive survey on transfer learning,'' \emph{Proceedings of the IEEE},
  vol. 109, no.~1, pp. 43--76, 2020.

\bibitem{ren2021survey}
P.~Ren, Y.~Xiao, X.~Chang, P.-Y. Huang, Z.~Li, B.~B. Gupta, X.~Chen, and
  X.~Wang, ``A survey of deep active learning,'' \emph{ACM computing surveys
  (CSUR)}, vol.~54, no.~9, pp. 1--40, 2021.

\bibitem{firal-neurips}
Y.~Chen and G.~Biros, ``Firal: An active learning algorithm for multinomial
  logistic regression,'' \emph{Advances in Neural Information Processing
  Systems}, vol.~36, 2024.

\bibitem{cupy17}
R.~Nishino and S.~H.~C. Loomis, ``{CuPy}: A numpy-compatible library for
  {NVIDIA} {GPU} calculations,'' \emph{31st conference on neural information
  processing systems}, vol. 151, no.~7, 2017.

\bibitem{gropp-mpi99}
W.~Gropp, E.~Lusk, and A.~Skjellum, \emph{Using MPI: portable parallel
  programming with the message-passing interface}.\hskip 1em plus 0.5em minus
  0.4em\relax MIT press, 1999, vol.~1.

\bibitem{dalcin2021mpi4py}
L.~Dalcin and Y.-L.~L. Fang, ``{mpi4py}: Status update after 12 years of
  development,'' \emph{Computing in Science \& Engineering}, vol.~23, no.~4,
  pp. 47--54, 2021.

\bibitem{Li2013}
X.~Li and Y.~Guo, ``Adaptive active learning for image classification,'' in
  \emph{2013 IEEE Conference on Computer Vision and Pattern Recognition}, 2013,
  pp. 859--866.

\bibitem{Sener2017}
O.~Sener and S.~Savarese, ``Active learning for convolutional neural networks:
  A core-set approach,'' \emph{arXiv preprint arXiv:1708.00489}, 2017.

\bibitem{Gissin2019discriminative}
D.~Gissin and S.~Shalev-Shwartz, ``Discriminative active learning,''
  \emph{arXiv preprint arXiv:1907.06347}, 2019.

\bibitem{citovsky2021batch}
G.~Citovsky, G.~DeSalvo, C.~Gentile, L.~Karydas, A.~Rajagopalan,
  A.~Rostamizadeh, and S.~Kumar, ``Batch active learning at scale,''
  \emph{Advances in Neural Information Processing Systems}, vol.~34, pp.
  11\,933--11\,944, 2021.

\bibitem{Gal2017}
Y.~Gal, R.~Islam, and Z.~Ghahramani, ``Deep bayesian active learning with image
  data,'' in \emph{International conference on machine learning}.\hskip 1em
  plus 0.5em minus 0.4em\relax PMLR, 2017, pp. 1183--1192.

\bibitem{Pinsler2019bayesian}
R.~Pinsler, J.~Gordon, E.~Nalisnick, and J.~M. Hern{\'a}ndez-Lobato, ``Bayesian
  batch active learning as sparse subset approximation,'' \emph{Advances in
  neural information processing systems}, vol.~32, 2019.

\bibitem{Hutchinson1990}
\BIBentryALTinterwordspacing
M.~Hutchinson, ``A stochastic estimator of the trace of the influence matrix
  for laplacian smoothing splines,'' \emph{Communications in Statistics -
  Simulation and Computation}, vol.~19, no.~2, pp. 433--450, 1990. [Online].
  Available: \url{https://doi.org/10.1080/03610919008812866}
\BIBentrySTDinterwordspacing

\bibitem{mvapich}
\BIBentryALTinterwordspacing
D.~K. Panda, H.~Subramoni, C.-H. Chu, and M.~Bayatpour, ``The mvapich project:
  Transforming research into high-performance mpi library for hpc community,''
  \emph{Journal of Computational Science}, vol.~52, p. 101208, 2021, case
  Studies in Translational Computer Science. [Online]. Available:
  \url{https://www.sciencedirect.com/science/article/pii/S1877750320305093}
\BIBentrySTDinterwordspacing

\bibitem{Rajeev2005}
\BIBentryALTinterwordspacing
R.~Thakur, R.~Rabenseifner, and W.~Gropp, ``Optimization of collective
  communication operations in mpich,'' \emph{Int. J. High Perform. Comput.
  Appl.}, vol.~19, no.~1, p. 49–66, feb 2005. [Online]. Available:
  \url{https://doi.org/10.1177/1094342005051521}
\BIBentrySTDinterwordspacing

\bibitem{deng-2012mnist}
L.~Deng, ``The mnist database of handwritten digit images for machine learning
  research,'' \emph{IEEE Signal Processing Magazine}, vol.~29, no.~6, pp.
  141--142, 2012.

\bibitem{cifar10}
\BIBentryALTinterwordspacing
A.~Krizhevsky, V.~Nair, and G.~Hinton, ``Cifar-10 (canadian institute for
  advanced research).'' [Online]. Available:
  \url{http://www.cs.toronto.edu/~kriz/cifar.html}
\BIBentrySTDinterwordspacing

\bibitem{caltech101}
F.-F. Li, M.~Andreeto, M.~Ranzato, and P.~Perona, ``Caltech 101,'' Apr 2022.

\bibitem{imagenet}
O.~Russakovsky, J.~Deng, H.~Su, J.~Krause, S.~Satheesh, S.~Ma, Z.~Huang,
  A.~Karpathy, A.~Khosla, M.~Bernstein, A.~C. Berg, and L.~Fei-Fei, ``{ImageNet
  Large Scale Visual Recognition Challenge},'' \emph{International Journal of
  Computer Vision (IJCV)}, vol. 115, no.~3, pp. 211--252, 2015.

\bibitem{dinov2}
M.~Oquab, T.~Darcet, T.~Moutakanni, H.~V. Vo, M.~Szafraniec, V.~Khalidov,
  P.~Fernandez, D.~Haziza, F.~Massa, A.~El-Nouby, R.~Howes, P.-Y. Huang, H.~Xu,
  V.~Sharma, S.-W. Li, W.~Galuba, M.~Rabbat, M.~Assran, N.~Ballas, G.~Synnaeve,
  I.~Misra, H.~Jegou, J.~Mairal, P.~Labatut, A.~Joulin, and P.~Bojanowski,
  ``Dinov2: Learning robust visual features without supervision,'' 2023.

\bibitem{scikit-learn}
F.~Pedregosa, G.~Varoquaux, A.~Gramfort, V.~Michel, B.~Thirion, O.~Grisel,
  M.~Blondel, P.~Prettenhofer, R.~Weiss, V.~Dubourg, J.~Vanderplas, A.~Passos,
  D.~Cournapeau, M.~Brucher, M.~Perrot, and E.~Duchesnay, ``Scikit-learn:
  Machine learning in {P}ython,'' \emph{Journal of Machine Learning Research},
  vol.~12, pp. 2825--2830, 2011.

\bibitem{nvidia2020}
Nvidia, ``Nvidia a100 tensor core gpu architecture,''
  \url{https://www.nvidia.com/content/dam/en-zz/Solutions/Data-Center/nvidia-ampere-architecture-whitepaper.pdf},
  2020, accessed: Apr. 2, 2024.

\end{thebibliography}

\end{document}